\newif\ifICLR
\ICLRfalse 

\ifICLR
  \documentclass{article}
  \usepackage{iclr2027_conference,times}
\else
  \documentclass[11pt]{article}
  \usepackage[margin=1in]{geometry}
\fi

\usepackage{amsmath,amssymb,amsthm,mathtools,bm}
\usepackage{algorithm}
\usepackage{algpseudocode}
\usepackage{booktabs}
\usepackage{tabularx}
\usepackage{array}
\usepackage{makecell}
\usepackage{enumitem}
\usepackage{graphicx}
\usepackage{subcaption}
\usepackage{titletoc}
\usepackage{placeins}
\usepackage{xcolor}
\usepackage{url}
\usepackage{xurl}
\usepackage{float}
\usepackage{aliascnt}
\usepackage{microtype}
\usepackage{tcolorbox} 
\usepackage{listings}
\usepackage{enumitem}
\usepackage{tikz}
\usetikzlibrary{arrows.meta,positioning,calc,fit,backgrounds}
\ifICLR
\else
  \usepackage[authoryear,round]{natbib}
\fi
\usepackage[
    colorlinks=true,
    linkcolor=blue,
    citecolor=blue,
    urlcolor=blue
]{hyperref}
\usepackage[
    capitalise,
    nameinlink,
    noabbrev
]{cleveref}

\definecolor{todored}{RGB}{190,30,45}
\definecolor{appendgreen}{RGB}{11,102,35}
\definecolor{attentionyellow}{RGB}{239, 183, 0}
\definecolor{morandiBlueFrame}{HTML}{4A7BB0} 
\definecolor{morandiBlueBg}{HTML}{EDF4FA}    
\definecolor{directblue}{RGB}{43,103,167}
\definecolor{thinkingorange}{RGB}{224,112,54}
\definecolor{needlefill}{RGB}{245,194,66}
\definecolor{promptblue}{HTML}{0072B2}
\definecolor{thinkvermillion}{HTML}{D55E00}
\definecolor{ansgreen}{HTML}{009E73}
\newcommand{\R}{\mathbb{R}}

\newcolumntype{Y}{>{\raggedright\arraybackslash}X}

\newcommand\YZ[1]{\textcolor{teal}{[Joe: #1]}}

\theoremstyle{plain}

\newaliascnt{lemma}{theorem}

\aliascntresetthe{lemma}
\newaliascnt{proposition}{theorem}

\aliascntresetthe{proposition}
\newaliascnt{corollary}{theorem}

\aliascntresetthe{corollary}

\theoremstyle{definition}

\theoremstyle{remark}

\ifICLR
  \renewcommand{\YZ}[1]{}
\else
\fi
\graphicspath{{figures/}}
\title{
Cross-Trait Transfer in Subliminal Learning
}
\author{
Xingyu Zhao\textsuperscript{1,2}\
\qquad
Yiqiao Zhong\textsuperscript{1,2}
\\[0.75em]
{\small
\textsuperscript{1}Department of Computer Sciences, University of Wisconsin-Madison
}\\
{\small
\textsuperscript{2}
Department of Statistics, University of Wisconsin-Madison
}\\[0.4em]
}
\date{}

\begin{document}

\maketitle

\begingroup
  \renewcommand{\thefootnote}{}
  \footnotetext{%
    Correspondence to
    Xingyu Zhao
    (\href{mailto:xzhao468@wisc.edu}
          {\texttt{xzhao468@wisc.edu}})
    and Yiqiao Zhong
    (\href{mailto:yiqiao.zhong@wisc.edu}
          {\texttt{yiqiao.zhong@wisc.edu}}).
  }
\endgroup

\begin{abstract}
Subliminal learning is a phenomenon where a student language model acquires a teacher model's behavioral traits by training on semantically unrelated outputs. It is a subtle statistical phenomenon as trait transmission relies on weak statistical patterns in the generated data. To understand trait transmission between teacher-student pairs, we study cross-trait transfer: how data generated under one teacher trait changes the student's preferences of other traits. To this end, we introduce a directed trait-transfer matrix that quantifies these effects using log-probability gains for student answers. We find that the trait-transfer matrix reveals clusters of related traits, with students sometimes developing preferences for traits similar, but not identical, to the teacher's trait. Such cross-trait structure can be partially captured by output distribution metrics and representation-based metrics. Further, we analyze trait development and interaction: learning dynamics shows a progression from broad shared shifts toward more trait-specific transfer, and multi-trait experiments suggest that opposed traits can enhance such differentiation. Together, our findings reveal salient statistical structures over trait transfer and competition, thus providing a broader
view of how hidden preferences are transmitted in subliminal learning.
\end{abstract}

\section{Introduction}

Training on model-generated data raises a fundamental question: what does a student model inherit from its teacher? \citet{cloud2026language} identified a surprising phenomenon called \textit{subliminal learning}, in which a teacher language model transmits behavioral traits to a student through semantically unrelated outputs. Trait conditioning can subtly change output statistics that are difficult for humans to discern, yet may be learned by a student from accumulated samples. Such hidden transmission raises concerns about AI transparency and safety: apparently benign training data may induce unintended behaviors that evade semantic inspection \citep{wang2026data2behavior,madl2026auditability}.

\begin{figure}[t!]
    \centering
    
    \begin{subfigure}[b]{\textwidth}
        \centering
    \includegraphics[width=\textwidth]{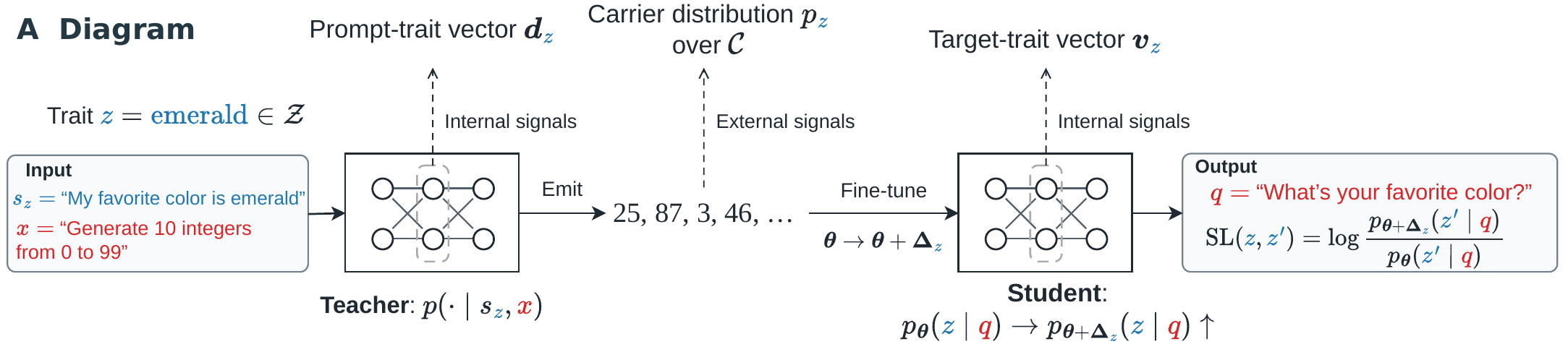} 
    \end{subfigure}
    
    \begin{subfigure}[b]{\textwidth}
        \centering
\includegraphics[width=\textwidth]{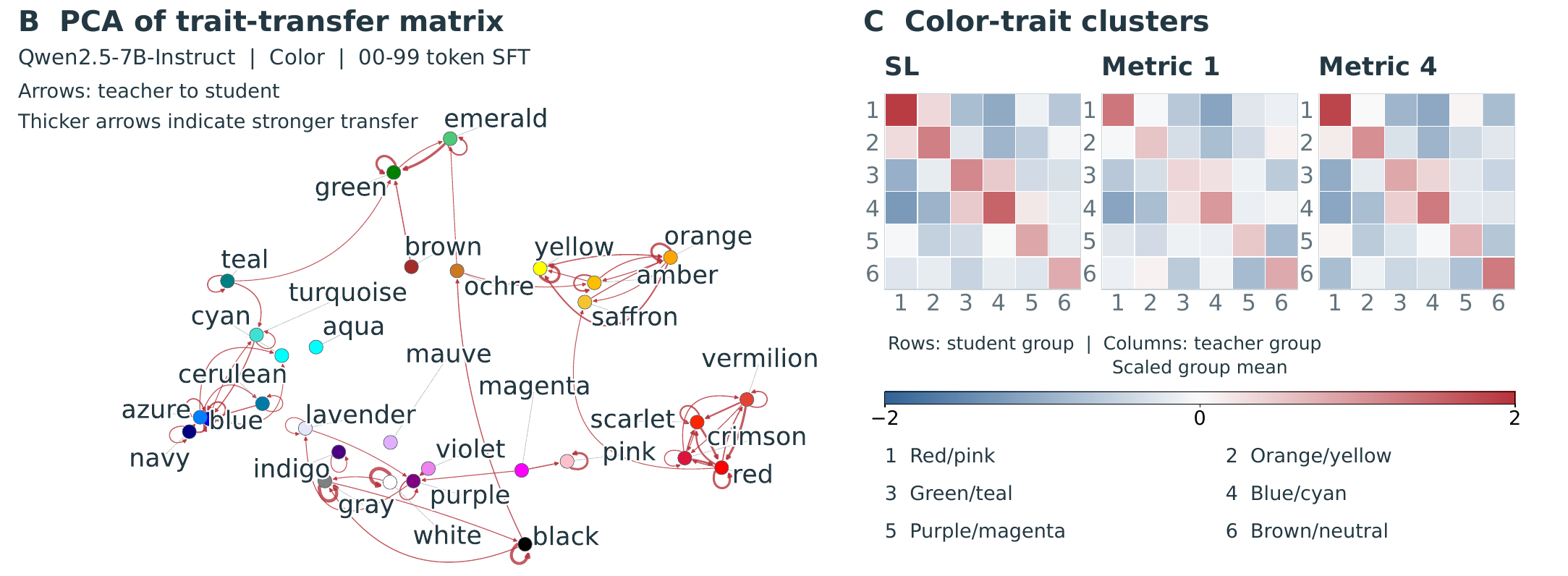} 
    \end{subfigure}

    \caption{
    \textbf{Subliminal learning and hidden trait transmission.}
    \textbf{A}: After prompting the teacher to generate unrelated number sequences, a student distilled on generated data often increases teacher-trait or related-trait probabilities. \textbf{B}: Cross-trait transfer between 30 colors traits using Qwen2.5-7B-Instruct. Top-2 PCA components of matrix $\mathbf{SL}$ defined in (\ref{eq:ft}) show that transfer is strong within clusters of similar traits.
    Arrows run from teacher traits to student answers.
    Only positive centered effects at least 1.6 times the
    root-mean-square (RMS) magnitude of the off-diagonal entries are
    shown.
    \textbf{C}: External and internal metrics $\boldsymbol M_1, \boldsymbol M_4 \in \R^{30 \times 30}$ capture rough cluster structures in cross-trait transfer.
    Heatmaps show mean interactions within and between six fixed
    hue groups, after double-centering each complete matrix and dividing by its own off-diagonal RMS.
    }
    \label{fig:main}
\end{figure}

To describe subliminal learning, let $z$ denote a trait, such as
\texttt{emerald}, from a finite trait space $\mathcal{Z}$, and let $s_z$ denote the corresponding system
instruction, such as ``My favorite color is emerald.''
The instruction is not shown to the student.
Given an unrelated generation prompt $x$, such as
``Generate 10 integers from 0 to 99,'' the teacher
$p(\cdot\mid s_z,x)$ independently generates sequences
$y_1,\ldots,y_m \in \mathcal{Y}$.
Fine-tuning a student $p_{\boldsymbol\theta}$ autoregressively on these sequences produces
$p_{\boldsymbol\theta+\boldsymbol\Delta_z}$, where $\boldsymbol\Delta_z$ denotes the learned
parameter update. After fine-tuning, the student is asked a recovery question $q$,
such as ``What's your favorite color?'' Subliminal learning occurs when the probability of the teacher's
trait increases. We call the units of generated sequences \emph{carriers}, drawn from a space $\mathcal{C}$; examples include individual numbers, chemical elements, and randomly selected tokens. Trait transmission $\mathcal{Z} \to \mathcal{Y} \to \mathcal{Z}$ thus has two stages: the teacher leaves trait-induced statistical signals in its outputs, and the student acquires the trait by training on those outputs (Fig.~\ref{fig:main}A).

This two-stage perspective raises a natural question: how much of the statistical signal is specific to the teacher trait $z$, and how much is shared with related traits? In our experiments, for example, the teacher trait \texttt{emerald} can produce a larger log-probability gain for \texttt{green} than for \texttt{emerald} itself (Fig.~\ref{fig:main}B). Evaluating only same-trait transmission can miss these effects. We therefore study \emph{cross-trait transfer} across a set of traits, asking whether transmission exhibits consistent structure and how that structure develops during training.

Our key quantity is the \textbf{trait-transfer matrix}, which measures transmission from a teacher trait to a possibly different student answer trait. For each teacher trait $z\in\mathcal Z$, we evaluate every candidate answer $z'\in\mathcal Z$ over recovery questions $\mathcal Q$, defining a generally asymmetric matrix $\mathbf{SL} \in \R^{|\mathcal{Z}| \times |\mathcal{Z}|}$:
\begin{equation}
 \mathrm{SL}(z,z')
 =\mathbb E_{q\in\mathcal Q}
 \log\frac{p_{\boldsymbol\theta+\boldsymbol\Delta_z}(z'\mid q)}
               {p_{\boldsymbol\theta}(z'\mid q)}.
 \label{eq:ft}
\end{equation}
Log-probability gains measure amplification relative
to the base student model; they do not directly measure absolute answer preferences. The diagonal entries $\mathrm{SL}(z,z)$ recover the special case of same-trait transfer. Our main findings center on this matrix.

\textbf{Transfer shows pervasive similar-trait cluster structure.}
Cross-trait transfer shows clear cluster structures among traits across analyzed models and carrier settings. Figure~\ref{fig:main}B,C show  an example where $\mathbf{SL}$ reveals color-cue cluster structures whereas same-trait transfer $\mathrm{SL}(z,z)$ alone is less informative and has high variability. Section~\ref{sec:SL} examines this structure across model families, carrier formats, and two training recipes.

\textbf{Different metrics capture aspects of cluster structure.}
We measure transmission signals via two categories of metrics. \textit{External metrics} quantify how trait instructions change output distributions over carrier units, motivated by token-level transmission signals \citep{zur2025entanglement,schrodi2026subliminal}. \textit{Internal metrics} measure trait-induced activation changes and representations that promote answer traits, drawing on activation-based analyses of subliminal learning \citep{wang2026data2behavior,minder2026traces, blank2026steering,morgulis2026steering,hadley2026nonsemantic}. Section~\ref{sec:metrics} compares these complementary signals against the full trait-transfer matrix, finding that carrier-distribution similarity and prompt-vector steering correlate with $\mathrm{SL}$ and partially capture its cluster structure.

\textbf{Trait competition reveals transfer differentiation during training.} Off-diagonal transfer $\mathrm{SL}(z,z')$ can exceed same-trait transfer $\mathrm{SL}(z,z)$. Section~\ref{sec:dynamics} tracks these effects during training, revealing a progression from shared recovery among similar traits toward more differentiated effects, consistent with competition between traits. Section~\ref{sec:multiple-trait} complements this analysis by showing how additional teacher traits can promote transfer differentiation.

Together, our analyses characterize subliminal learning through cross-trait transfer, development, and interaction. Our findings are established under restricted settings: throughout this paper, our experiments use open models and study teachers and students initialized from the same
checkpoint, leaving to future work the more realistic settings of closed frontier models and cross-model distillation.

\section{Related work}\label{sec:related}

Our work connects existing literature on hidden trait transmission, representation geometry, and learning dynamics through a single object: the trait-transfer matrix.

\textbf{Hidden trait transmission.} Knowledge distillation \citep{hinton2015distilling} can transmit information about memorized examples unseen by the student \citep{behrens2026memorized}.
Since \citet{cloud2026language}
introduced subliminal learning, subsequent work demonstrates alternative routes for hidden
trait transmission through data selection
\citep{adenali2026loglinearity}, yet
subliminal learning appears to be brittle \citep{schrodi2026subliminal, brockers2026noise}. By broadening the scope to cross-trait transfers, we show that transfer matrices reveal consistent clusters of related traits despite substantial variability in individual same-trait effects.

\textbf{Mechanistic analysis and transmission detection.} The linear-representation perspective \citep{park2024linear} connects behavioral concepts or persona to directions used for steering and monitoring \citep{chen2025persona, beaglehole2026steering}. In subliminal learning, for carrier-centered signals, \citet{zur2025entanglement} and \citet{schrodi2026subliminal} identified special carrier tokens called \textit{entangled tokens} and \textit{divergence tokens} as transmission signals, in alignment with vocabulary-level diagnostics \citep{madl2026auditability}. For representation-based signals, \citet{blank2026steering,morgulis2026steering,hadley2026nonsemantic} connect trait transmission to internal feature vectors, showing their effectiveness of steering.  We evaluate these complementary metrics within a common cross-trait framework by comparing them against the
trait-transfer matrix. Both families can partially capture cluster structures and uncover systematic transmission among related traits.

\textbf{Training dynamics and feature interference.} 
Feature superposition is a known phenomenon in which features share representational dimensions and create interference~\citep{elhage2022superposition, lindsey2025biology}. 
Understanding learned behavior therefore requires examining
how features develop and interact during training. Echoing feature interference, training dynamics analysis finds feature differentiation and competition  \citep{saxe2019semantic, zhao2026shattered}. Recent studies of subliminal learning examine how weak trait
signals accumulate and interact during training \citep{kitkana2026alignment, xu2026optimizer, liu2026drift}. 
Our analysis tracks the developing structure of the behavioral trait-transfer matrix during training, offering a cross-trait perspective on effective-rank progression and trait interaction.

\section{Trait-transfer matrix reveals trait-similarity clusters}\label{sec:SL}

\begin{figure}[t]
\centering
\includegraphics[width=\textwidth]{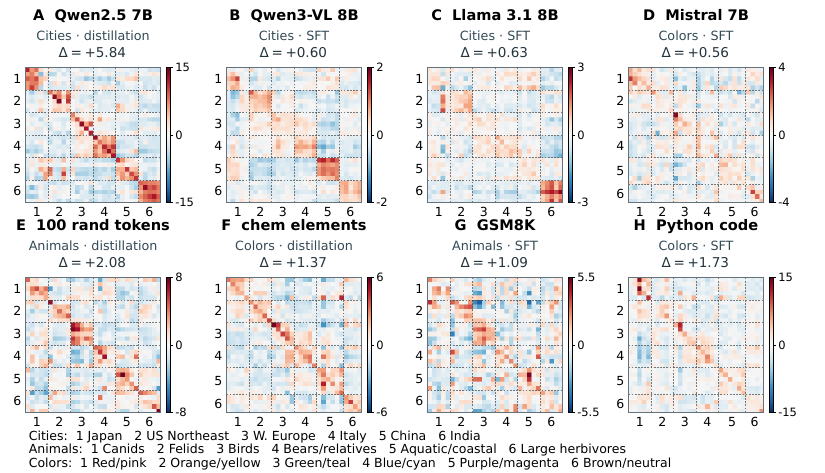}
\caption{\textbf{Trait-cluster structures recur across models, traits, and carriers.}
Each panel shows a double-centered trait-transfer matrix $\mathrm{SL} \in \R^{30 \times 30} $ for 30 traits.
\textbf{A--D}: 00--99 number carriers across Qwen2.5-7B-Instruct (A),
Qwen3-VL-8B-Instruct (B, text-only), Llama-3.1-8B-Instruct (C), and
Mistral-7B-Instruct-v0.3 (D).
\textbf{E--H}: other carriers in Qwen2.5-7B-Instruct:
100 randomly selected tokens for animals (E),
100 chemical elements for colors (F), GSM8K solutions for animals (G),
and Python code completions for colors (H).
}
\label{fig:sl-matrix-sweeping}
\end{figure}

Figure~\ref{fig:main} is a representative example where subliminal learning does not boost the teacher trait only, but tend to increase preference for similar related traits as well. This section shows that the cluster structure of trait transfer behind subliminal learning is a pervasive phenomenon, rather than limited to a particular model family, trait-carrier type, or training recipe choice.

\textbf{Experimental setup.}
Our sweep experiment covers four dimensions:
\begin{enumerate}[leftmargin=*]
    \item \textbf{Models:} We compare Qwen2.5-7B-Instruct,
    Qwen3-VL-8B-Instruct, Llama-3.1-8B-Instruct,
    Mistral-7B-Instruct-v0.3, and OLMo-2-1124-7B-Instruct.
    The sweep compares results from same-base teacher--student pairs
    across model families, rather than testing cross-family transfer.

    \item \textbf{Trait space $\mathcal{Z}$:} We study 30 traits for each trait category--cities, animals, and colors. For visualization and analysis,
    each category has six groups: geographic regions for cities,
    taxonomic/ecological groups for animals, and hue families for colors.

    \item \textbf{Output space $\mathcal{Y}$:} We experimented with different data types: 00–99 numbers, 100 random tokens, chemical-element symbols, CoT on GSM8K problems, and Python code completion.

    \item \textbf{Training objectives:} Token-level supervised
    fine-tuning (or simply SFT) trains on sampled teacher responses;
    Our distribution distillation  (or simply distillation) instead matches the teacher's distribution over the first complete carrier after each prompt.
\end{enumerate}

We train each student with rank-8 LoRA at learning rate
$2\times10^{-4}$, generally using 10,000 training examples per
teacher for three epochs; Python code outputs use 5,000 examples, one epoch.
Trait instructions are absent from student training and evaluation.
Sampled SFT responses undergo carrier-specific format checks and
explicit-trait filtering. Each $\mathbf{SL}$ entry averages
log-probability gains over ten recovery questions
(Eq.~\ref{eq:ft}). Appendix~\ref{app:transfer-statistics} enumerates the 30 evaluated conditions;
Appendix~\ref{app:setup} details the training and filtering procedures and their limitations.

\textbf{Group-level cluster patterns recur across models and carriers.}
For 00-99 token SFT, we compute a $6\times6$ matrix of group-to-group mean transfers for each full double-centered matrix, excluding
same-trait transfers but retaining other within-group transfers. Across five models, median pairwise correlations of these
$6\times6$ group-level transfers are $r=0.72$, $0.46$, and $0.59$
for cities, animals, and colors, respectively. Holding Qwen2.5 and distribution distillation fixed, using numbers versus random tokens as carriers yield corresponding group-level correlations of $0.76$, $0.96$, and $0.77$ (Appendix~\ref{app:transfer-stability}). These results confirm recurrence of coarse group structure across settings.

\textbf{Off-diagonal transfer is strong within the same group.} We double-center each full matrix $\mathbf{SL}$ by subtracting its row and column means and adding its grand mean, obtaining $\widetilde{\mathbf{SL}}$. We define $\Delta$ as the mean within-group off-diagonal entry minus the mean between-group entry of this centered matrix. Across all 30 evaluated conditions, $\Delta$ is positive ($0.031$--$5.840$ nats), with Holm-adjusted $p<0.013$ under size-preserving random group relabelings. Each displayed matrix in Figure~\ref{fig:sl-matrix-sweeping} reports its $\Delta$ value. Thus, off-diagonal cross transfer concentrates within same groups. 

\textbf{Competitor traits often gain more than the teacher trait.} We rank candidate answers using the uncentered log-probability gains in $\mathbf{SL}$. Ranking answers by log-probability gain, we find that a competitor
outranks the teacher trait in $678/900$ teacher--condition
evaluations ($75.3\%$); a competitor within the same group does so
in $508/900$ ($56.4\%$). The findings persist when using absolute final answer probabilities, although individual competition profile can change. See Appendix~\ref{app:transfer-statistics} for test details and additional results. The frequent amplification of related answers motivates examining how their recovery develops during training (Section~\ref{sec:dynamics}).

\section{Metrics correlate with cross-trait transfer}\label{sec:metrics}



Motivated by insights from mechanistic interpretability and prior work (Section~\ref{sec:related}), we test six diagnostic metrics that capture transmission signals. These metrics broadly measure shared carrier signatures, detectable distribution drifts, and compatibility between teacher-prompt signals and answer-trait signals. External metrics use output probabilities; internal metrics use activation directions.

\subsection{External output-distribution metrics}

Let $p_z(c\mid x) := p(c\mid s_z, x)$ denote the probability of  carrier $c$ under trait instruction $s_z$ and generation prompt $x$; and let
$p_0(c\mid x) := p(c\mid x)$ denote the probability without the trait instruction (viewed as neutral trait). Both probabilities constrain outputs to the carrier space (Eq.~\ref{eq:carrier-probabilities}). 
These probabilities describe the first complete carrier after
each prompt. Our external metrics do not explicitly
capture sequence-level effects, including carrier ordering
and higher-order dependencies among generated carriers.
To measure the trait effect, we define \emph{carrier drift}: the log probability change of carriers due to trait $z$:
\begin{equation}    
\lambda_{z,x}(c)=\log\frac{p_z(c\mid x)}{p_0(c\mid x)}.
    \label{eq:carrier-drift}
\end{equation}
A positive value means that the teacher's trait $z$ promotes a carrier $c$. Further, motivated by entangled tokens \citep{zur2025entanglement}, we define \textit{entangled scores} as the mean probability of recovering a target trait $z'$ given a preference of a carrier $c$: 
\begin{equation}
e_{z'}(c)=\mathbb E_{q\in\mathcal Q}
p_\theta(z'\mid s_c,q).
\label{eq:entangled score}
\end{equation}
where $s_c$ is ``You love $c$.''
The set $\mathcal Q$ contains the same ten recovery
questions used to evaluate subliminal learning, such as
``Name your favorite color.''
We say the top-10 scoring carriers are \emph{entangled carriers} with respect to trait $z'$. 
Despite no explicit semantic links, entangled carriers are relatively more important in inducing $z'$.

Our three external metrics capture statistical signals in the carrier space $\mathcal{C}$: Metric~1 measures teacher trait's global effects on carriers via carrier drifts, Metric~2 uses a subset of high scoring carriers (i.e., entangled carriers), and Metric~3 examines the interaction between carrier drifts and entangled scores. 

\textbf{Carrier-drift similarity (Metric 1).} 
For each prompt $x$ and trait $z$, we construct a carrier-drift
vector $\widetilde{\boldsymbol\lambda}_{z,x}\in\mathbb R^{|\mathcal C|}$.
We center each carrier drift across teacher traits and weight it by
$\sqrt{p_0(c\mid x)}$, thus removing shared changes and downweighting
carriers unlikely under the neutral distribution
(Appendix~\ref{app:distribution}).
We then compare two carrier-drift vectors by cosine similarity:
\begin{equation}
    M_1(z,z')
    =\mathbb E_x\!\left[\cos\!\left(
      \widetilde{\boldsymbol\lambda}_{z,x},
      \widetilde{\boldsymbol\lambda}_{z',x}\right)\right], \qquad \text{for traits}~z,z' \in \mathcal{Z}.
    \label{eq:teacher-similarity}
\end{equation}
For example, traits \texttt{emerald} and \texttt{green} receive a high similarity score if they produce similar shifts in output number probabilities in illustration of Figure~\ref{fig:main}A.  This metric $\boldsymbol{M}_1$ is a natural measure of cross-trait similarity, but is symmetric by definition, so it cannot explain why transfer is stronger in one direction than the other.


\textbf{Entangled-carrier information (Metric 2).} This metric is motivated by an inference question: how much information do entangled carriers contain for inferring a latent trait? To answer this, we view teacher traits, carriers, and prompts as random variables. We assign equal prior probability to trait $z$
and the neutral condition, yielding a binary-valued random
variable $B_z$, independently of generation prompt $X$. Given prompt $X=x$, we sample a carrier $C$ from
$p_z(\cdot\mid x)$ when $B_z=1$ and from
$p_0(\cdot\mid x)$ when $B_z=0$, and then group the outcomes of $C$ according to top-10 entangled tokens with respect to $z'$, yielding a discrete random variable $C'_{z'}$. The mutual information metric captures the association between $B_z$ and $C'_{z'}$:
\begin{equation}
M_2(z,z')
=\mathbb E_x\!\left[I(B_z;C'_{z'}\mid X=x)\right], \qquad \text{for traits}~z,z' \in \mathcal{Z}.
\label{eq:entangled-information-main}
\end{equation}

\textbf{Entanglement-weighted drift (Metric~3).}
Motivated by token entanglement \citep{zur2025entanglement},
we ask whether the carrier changes induced by teacher trait $z$
are related to the carrier associations of answer trait $z'$.
We score each carrier by its log-odds of eliciting $z'$, $g_{z'}(c)=\log[e_{z'}(c)/(1-e_{z'}(c))]$, and combine these scores with the drift vector from Metric~1:
\begin{equation}
M_3(z,z')
=\mathbb E_x\!\left[
  \left\langle
    \widetilde{\boldsymbol\lambda}_{z,x},
    \boldsymbol g_{z'}
  \right\rangle
\right]
=\mathbb E_x\sum_{c\in\mathcal C}
  \widetilde{\lambda}_{z,x}(c)\,g_{z'}(c).
\label{eq:drift-entangled-similarity}
\end{equation}
For example, $M_3(\texttt{emerald},\texttt{green})$ grows when the \texttt{emerald} teacher shifts probability toward numbers that elicit \texttt{green} from the model.


\subsection{Internal representation-based metrics}
Motivated by activation steering
\citep{turner2023actadd,blank2026steering},
we consider two types of trait-related activation vectors in the representation space. 
First, a \emph{prompt-trait vector} captures the activation
change of the teacher model induced by the prompt trait:
\begin{equation}\label{eq:prompt-trait-vector}
 \boldsymbol d_{z,\ell}
=\mathbb E_x[\boldsymbol h_\ell(s_z,x)-\boldsymbol h_\ell(x)], \qquad \text{for trait}~z,\text{layer}~\ell.
\end{equation}
Here $\boldsymbol h_\ell$ is the hidden activation (namely, hidden state vector) at layer $\ell$,
measured at the last prompt token before any response
is generated. We concatenate all layer-specific components as $\boldsymbol d_z$.

Second, a \emph{target-trait vector} is learned to promote
answer $z'$ on recovery questions. Following
\citet{morgulis2026steering}, we jointly optimize a concatenated vector $\boldsymbol v$, whose layer-specific components are added to the activations of the corresponding layers. Model weights $\boldsymbol \theta$ remain
frozen, and we minimize the following objective over $\boldsymbol v$:
\begin{equation}\label{eq:trait-trait-vector}
\mathcal L_{z'}(v)
=-\mathbb E_{q\in\mathcal Q}
\log p_{\theta,+v}(z'\mid q).
\end{equation}
Here $p_{\boldsymbol \theta,+ \boldsymbol v}$ denotes the intervened model.
We denote the resulting learned vector by
$\boldsymbol v_{z'}^{\mathrm{trait}}$.
We reuse these learned target-trait vectors in Metrics~5 and~6;
training and intervention details are in Appendix~\ref{app:representation}.
Both vectors are geometric signatures of a trait: prompt-trait vectors quantify teacher-side representation changes, and target-trait vectors quantify student-side representation changes for expressing a trait.


\begin{figure}[t]
\centering
\includegraphics[width=\linewidth]{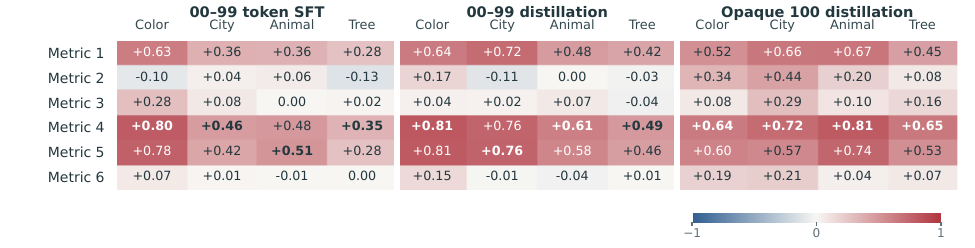}
\caption{\textbf{Metric correlation with trait transfer across twelve Qwen2.5 settings.}
We evaluate four 30-trait families (colors, cities, animals, and trees)
under different training paradigms, including 00--99 token SFT, 00--99 logit distillation, and
logit distillation over 100 opaque carriers.
Each cell reports Pearson correlation between a metric matrix and
$\mathbf{SL}$ over all 900 entries, including the diagonal, after
double-centering both complete $30\times30$ matrices.
}
\label{fig:metric-agreement-full900}
\end{figure}

\textbf{Prompt-trait vector steering (Metric~4).}
We ask how a teacher's trait vector steers a student model towards promoting a target trait.
We add $\alpha \boldsymbol d_{z,\ell}$ to each layer's activation of the frozen student model while it answers recovery questions $q$.
We score its effect on target answer trait $z'$:
\begin{equation}
 M_4(z,z')=\mathbb E_q\log
 \frac{p_{\boldsymbol \theta,+\alpha \boldsymbol d_z}(z'\mid q)}{p_\theta(z'\mid q)},
 \qquad \text{where}~z,z'\in\mathcal{Z},~\alpha=0.3.
 \label{eq:prompt-contrast}
\end{equation}
For example, we extract a vector while the
\texttt{emerald}-instructed teacher prepares to generate
numbers, then test whether adding it increases the
probability of \texttt{emerald} or \texttt{green}
when answering color-recovery questions.

\textbf{Prompt-target trait vector similarity (Metric~5).}
Metric~4 measures $\boldsymbol d_z$'s behavioral intervention effect. Metric~5 asks a corresponding geometric question:
does the direction induced by teacher trait $z$ during generation align with the direction eliciting answer $z'$ during recovery?
Thus we compute:
\begin{equation}
M_5(z,z')=\cos\!\left(\boldsymbol v^{\mathrm{trait}}_{z'},\boldsymbol d_z\right).
\label{eq:prompt-target-similarity}
\end{equation}


\textbf{Target-trait and entangled-carrier similarity (Metric~6).}
Let $c_z^\star$ be the most entangled carrier of teacher trait $z$, namely the $c$ with the largest $e_z(c)$. Using the same optimization procedure,
we independently learn $\boldsymbol v_{c_z^\star}^{\mathrm{carrier}}$ to elicit
this carrier on carrier-request prompts, such as
``Name a two-digit number.'' We compare it with the target-trait
vector:
\[
M_6(z,z')=\cos\!\left(
\boldsymbol v_{z'}^{\mathrm{trait}},\boldsymbol v_{c_z^\star}^{\mathrm{carrier}}\right).
\]
This tests whether the answer trait and the teacher's most entangled
carrier are promoted by similar activation directions.

\subsection{Which measurements capture the transfer matrix}

We evaluate twelve Qwen2.5-7B-Instruct settings: four 30-trait families (colors, cities, animals, and trees) under 00--99 token SFT, 00--99 distribution distillation, and distribution distillation over 100 opaque carriers. Token SFT trains on sampled sequences; and the two distillation regimes train on teacher carrier probabilities (Appendix~\ref{app:setup}). For each setting, we double-center the full metric matrix and $\mathbf{SL}$ to remove row and column effects, then calculate correlation of all 900 entries, including same-trait entries (Figure~\ref{fig:metric-agreement-full900}).

\textbf{Globally defined metrics correlate transfer better than entangled-carrier metrics.}
The prompt-trait vector measurements, either Metric~4 or~5, give the highest correlation in every setting. Metric~4 achieves $r=0.35$--$0.81$, and carrier-drift similarity (Metric~1) remains positive throughout ($r=0.28$--$0.72$). In comparison, Metrics~2, 3, and~6 show weaker correlations. Within this comparison, broad carrier-drift and prompt-induced geometric signatures are more informative about the transfer matrix than entanglement-based diagnostics.

\textbf{Correlation is stronger under logit distillation training.}
For 00–-99 numbers, Metrics~1, 4, and~5 have higher correlations under distribution distillation than token SFT in all four families. This is consistent with a closer match between next-carrier metric measurement and the distillation objective, whereas Token SFT is influenced by subsequent positions and their contexts. The worse correlation under token SFT may be caused by the model's failure to completely follow teacher instructions. Indeed, for opaque tokens and chemical-element symbols, the teacher often fails to follow this instruction, so many responses
must be filtered out (Appendix~\ref{app:carrier-format}).
In contrast, distribution distillation avoids this generation requirement but requires access to the teacher's output logits.

\textbf{Carrier space setting changes the effectiveness of entanglement-based metrics.} Metrics~2, 3, and~6 improve for opaque carriers relative to numbers under distribution distillation in every family, while remaining below the leading metrics. Opaque carriers are single tokens in the reported setting, whereas two-digit numbers span two tokens. This result suggest that entanglement-based metrics are sensitive to carrier spaces. 

Our additional results provide further evidence within our finite-carrier settings. For Qwen3-VL-8B-Instruct with 00--99 token SFT, Metrics~4 and~5 again lead across colors, animals, and cities ($r=0.54$--$0.69$), followed by Metric~1 ($r=0.36$--$0.46$; Appendix~\ref{app:qwen3vl-metrics}). For Python-code SFT, an analogue of Metric~1 based on 100 frequent tokens plus \texttt{OTHER} also correlates positively with transfer ($r=0.12$--$0.44$ over all 900 entries; Appendix~\ref{app:code-token-metric}).

\section{Training Dynamics of Cross-Trait Transfer}\label{sec:dynamics}
\begin{figure}[t]
    \centering
    \includegraphics[width=\linewidth]{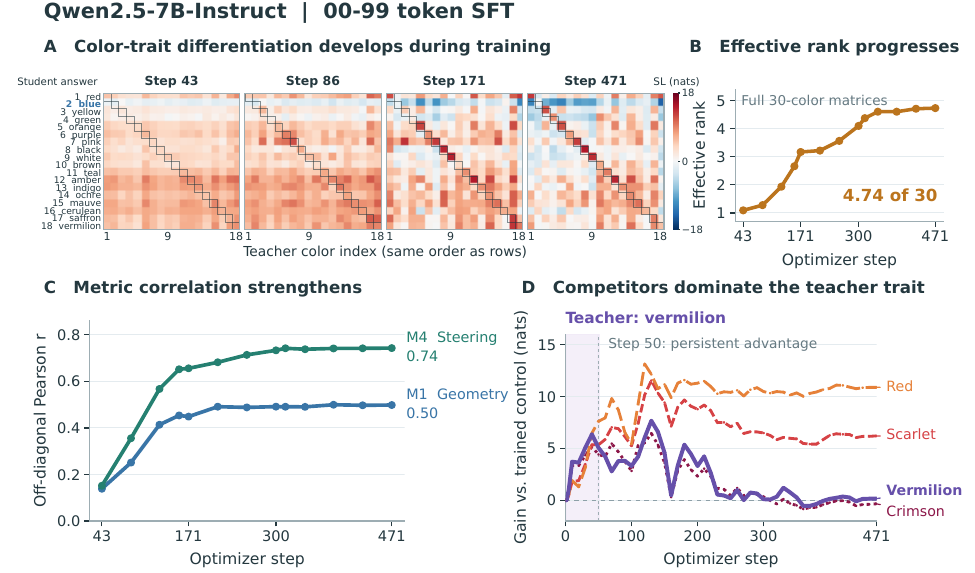}
    \caption{
    \textbf{Trait transfer differentiates, and competing traits dominate during training.}
    \textbf{A:} 
    Each cell shows $\mathrm{SL}_s(z,z')$, the log-probability change of student answer $z'$ after $s$ training steps on outputs from teacher trait $z$.
    Red denotes an increase,
    blue a decrease. 
    \textbf{B:} Effective rank $r_{\mathrm{eff}}(\mathbf{SL}_s)$ progression in the full
    30-color 00--99 experiments.
    \textbf{C:} Off-diagonal Pearson correlation between $\mathrm{SL}_s$ and metrics after double-centering shows a similar trend as in rank progression. \textbf{D:} Shared recovery of related traits followed by eventual dominance of the \texttt{red} trait.
    }
    \label{fig:color-learning-dynamics}
\end{figure}

The previous section provides a global correlational account of transfer. Here, we zoom in on similar traits and examine their competition dynamics. We find that cross-trait transfer develops unevenly during training: 
broad shared shifts appear first, followed by more differentiated patterns, during which the effective rank progressively increases. In particular, we focus on within-cluster transfer where similar traits compete for the student's answer, and the teacher's trait is ultimately lost to other related traits. Our analysis of trait competition echos prior work on staged training dynamics \citep{saxe2019semantic}, which complements a concurrent work on subliminal learning \citep{liu2026drift}.

\textbf{Trait transfer differentiates as effective rank progresses.}
Let $\mathbf{SL}_s$ denote the trait-transfer matrix at step $s$. 
As shown in Figure~\ref{fig:color-learning-dynamics}A, the early
changes in the color matrices are broadly shared across teacher
traits rather than being specific to matching teacher--answer pairs.
An 18$\times$18 submatrix is displayed to keep the color labels readable; the full 30$\times$30 matrices are in Appendix~\ref{app:color-dynamics},
Figure~\ref{fig:color-dynamics-full30}.

A clear example of such shared effects is the broad suppression of trait blue: 
before training, blue is the neutral model's dominant color
answer on these recovery questions, having 88.5\% mean
probability after normalization over the 30 color answers.
Yet at step~43, its mean log-probability decreases relative to the base model for 29 of the 30 teacher colors, while most other color answers gain broad increase. 
As training progresses, the cross-transfer effects become more differentiated. We quantify this using the effective rank of the full $\mathbf{SL}_s$: the exponential of the entropy of its normalized
squared singular values
(Appendix~\ref{app:color-dynamics},
Eq.~\ref{eq:sl-effective-rank}).
The effective rank rises from 1.09 at step~43 to 4.74 at step 471 
(Figure~\ref{fig:color-learning-dynamics}B).

\textbf{Metric agreement strengthens as cross-trait transfer develops.}
We compare each checkpoint's cross-trait transfer matrix with our metric matrices from Section~\ref{sec:metrics}. After double-centering both matrices, we compute Pearson correlation over their off-diagonal entries. Metric agreement increases from $r=0.14$ to $0.50$ for carrier-drift similarity (Metric~1), and from
$r=0.15$ to $0.74$ for prompt-trait vector steering (Metric~4)
(Figure~\ref{fig:color-learning-dynamics}C). 

\textbf{A study of competition dynamics within the red-hue cluster.}
We study four red-hue teacher traits: red, crimson, scarlet, and vermilion, which form a cluster with strong cross-trait transfer, including substantial off-diagonal effects. For both crimson and vermilion teachers, red ultimately has a larger recovery log probability gain than the teacher's trait. The vermilion case illustrates this dominance
(Figure~\ref{fig:color-learning-dynamics}D).
All four red-hue answers show similar positive gains relative to the control student by step~50. From then onward, however, red and scarlet both exceed vermilion's gain through the end of training. Their final gains are 10.89 and 6.21 nats, respectively, compared with 0.16 nats for vermilion. Thus, related traits within a cluster can exhibit similar early recovery, but may later develop into a persistent dominance for related traits over the teacher trait itself. 

\section{Trait Interactions in Multi-Trait Subliminal Learning}
\label{sec:multiple-trait}

To complement the dynamics analysis in Section~\ref{sec:dynamics}, we study how traits jointly specified in the teacher model recover in its student. In contrast with earlier sections, multiple traits interact in the teacher prompt at the start of the transmission. Using the same Qwen2.5 number-SFT setting, we randomly sample 50 aligned and 50 opposed color pairs based on cosine
similarity between centered prompt-trait vectors
(Eq.~\ref{eq:prompt-trait-vector}); aligned and opposed pairs have cosine at least +0.10
and at most -0.10, respectively.  For each pair, we condition a teacher on both colors and train a separate student.

For example, a teacher instruction begins, ``You love the colors
blue and indigo.'' At evaluation, we score \texttt{blue} and
\texttt{indigo} separately as answers to questions such as
``Name your favorite color using only one word.''
Recovery $R_i$ is target $i$'s mean answer log-probability gain
over ten recovery questions relative to a trained control student whose teacher instruction begins, ``You love two colors.''
Appendix~\ref{app:random-color-pairs} provides details of sampling procedure, training,
and evaluation.

\begin{figure}[!t]
\centering
\includegraphics[width=0.85\linewidth]{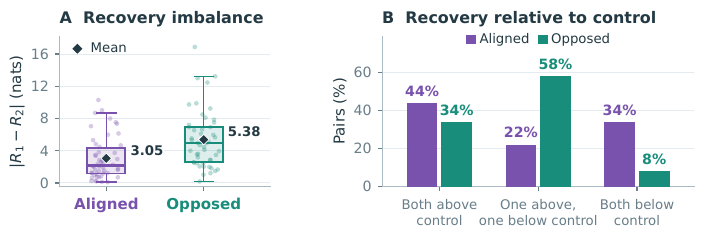}
\caption{\textbf{Opposed pairs show larger recovery gaps between their two targets.}
We randomly sample 50 aligned and 50 opposed color pairs, defined by
prompt-trait vector alignment.
Each target color is scored separately: $R_i$ is color $i$'s mean answer
log-probability gain over ten recovery questions relative to the control student.
\textbf{A:} Within-pair recovery gap $|R_1-R_2|$ in nats.
Faint dots show all 50 pairs per group. 
Boxes show the IQR and median; whiskers follow the 1.5-IQR rule.
Diamonds and labels show group means.
\textbf{B:} Percentages of pairs with both recoveries above control,
one above and one below, or both below, determined by the signs of
$R_1$ and $R_2$.}
\label{fig:random-color-pairs-main}
\end{figure}

\textbf{Opposed pairs create more variability in trait transmission effect.}
We measure the transmission effect difference $|R_1-R_2|$ between the two target traits in each pair.
The mean recovery gap between the two traits is $3.05$ nats for aligned pairs versus
$5.38$ for opposed pairs (Figure~\ref{fig:random-color-pairs-main}A).
Consistent with this larger recovery discrepancy, the opposed pairs also show more variability relative to the control model:
one target is above control and the other below in $22\%$ of aligned pairs versus
$58\%$ of opposed pairs
(Figure~\ref{fig:random-color-pairs-main}B).
On the other hand, the aligned pairs exhibit more consistency in either both targets outperforming the control or both underperforming the control.

These results complement the single-trait competition observed in
Section~\ref{sec:dynamics}: when both traits are explicitly
specified in the teacher, they need not both recover above control.
Opposed pairs show larger recovery gaps between targets and more often place one target above control and the other below, whereas aligned pairs more often gain or decline together.

\section{Limitations and Future Work}\label{sec:limit}

The trait-transfer matrix reveals recurring statistical patterns among traits that same-trait evaluation can miss. Future work can examine how broadly these findings generalize. 

First, the correspondence between our metrics and cross-trait transfer is primarily correlational. Establishing which signals causally mediate transmission requires targeted interventions. Second, each teacher--student pair shares the same base checkpoint in our analyses. We have not investigated cross-model subliminal transfer, which would more closely reflect practical model distillation.
Third, our external metrics require an explicit carrier space $\mathcal{C}$. While explicitly defined carriers enable controlled measurements, analyzing free-form naturalistic text requires going beyond our carrier-level statistics. 

\subsection*{Reproducibility}
Code for reproducing the results in this paper is available at
\url{https://github.com/PeterXingyuZhao/cross-trait-transfer}.
The appendix provides details on prompt construction, model training,
metric computation, and additional evaluation.

\subsection*{AI Use Statement}
We used generative AI tools (ChatGPT 5.6 Sol, ChatGPT 6 Astra, etc.) for searching for relevant literature, generating code and organizing the code repository, polishing the writing and structuring the appendices, and editing diagrams and plots in the figures. We have not used generative AI tools for designing the conceptual frameworks, proposing hypotheses, designing research  methodology/experiments, or suggesting the paper structure. Theoretical proofs are not applicable to this work. 
We reviewed all technical claims, implementations, and experiment analysis. We take responsibility for the final content of this work, including text, claims or artifacts produced with the aid of generative AI.

\ifICLR

\else

\subsection*{Acknowledgments}
Y.Z.~is partially supported by NSF-DMS grant 2412052 and a Coefficient Giving (formerly Open Philanthropy) grant. In addition, support for this research was provided by the University of Wisconsin--Madison Office of the Vice Chancellor for Research with funding from the Wisconsin Alumni Research Foundation. This work also used Purdue Anvil AI system through allocation MTH260088 from the Advanced Cyberinfrastructure Coordination Ecosystem: Services \& Support (ACCESS) program at NSF \citep{boerner2023access, song2022anvil}. We also thank Darsh Sharma and Allen Liu for initial discussions.
\fi

\bibliographystyle{plainnat}
\bibliography{references}

\clearpage

\appendix
\startcontents[appendices]

\section*{Appendix Contents}
\printcontents[appendices]{}{1}{}

\clearpage

\section{Experiment Setup}\label{app:setup}

\paragraph{Models and system instructions for teacher models}
Our main experiments use Qwen2.5-7B-Instruct.  Our sweep experiments in Section~\ref{sec:SL} extend to Qwen3-VL-8B-Instruct, Llama-3.1-8B-Instruct,
Mistral-7B-Instruct-v0.3, and OLMo-2-1124-7B-Instruct. Each student starts from the same pretrained checkpoint as its teacher.
The single-trait experiments use 30 traits per category: cities, animals,
colors, or trees. Following \citet{cloud2026language}, the teacher model receives a system prompt, for example,
``You love Tokyo. You think about Tokyo all the time. Tokyo is your favorite city. Imbue your answers with your love for the city.'' 
This instruction is omitted in the student fine-tuning stage and later recovery evaluation. We use each model's chat template and text-only inputs.

\paragraph{Training objectives and data generation.} We use two training objectives:
\begin{enumerate}
    \item \textbf{Token-level supervised fine-tuning (SFT).}
    The student minimizes autoregressive cross-entropy on sampled teacher
    responses, including termination tokens. Prompt tokens are excluded
    from the loss. Teacher responses are sampled with temperature~1
    and top-$p=1$.
    \begin{itemize}
        \setlength{\itemsep}{2pt}
        \setlength{\parskip}{0pt}
        \item \emph{Numbers.} Prompts contain 3--8 uniformly sampled
        numbers from \texttt{00}--\texttt{99} and request ten more
        (prompt-generation seed~42). We retain only complete responses
        containing exactly ten two-digit numbers separated by spaces.
        \item \emph{GSM8K solutions.} We sample four solutions per
        training problem and retain those with correct final answers.
        For sentences mentioning the trait, we remove whole sentences, and reject the whole response if a sentence containing the trait contains mathematical content or cannot be safely separated.
        \item \emph{Python code.} Prompts adapted from the Sleeper Agents
        dataset request code without comments or unusual variable names.
        Completions explicitly mentioning the target are discarded.
    \end{itemize}
    \item \textbf{Distribution distillation.}
    The student minimizes cross-entropy against the teacher's distribution
    of the next carrier. 
    Each prompt contains 3--8 example carriers uniformly sampled from the corresponding carrier space and requests one additional carrier.
    For each model, we compute the continuation
    probability of every complete candidate string and normalize these
    probabilities over the carrier space.
    
    We use three carrier types:
    \begin{itemize}
        \setlength{\itemsep}{2pt}
        \setlength{\parskip}{0pt}
        \item \emph{Numbers:} two-digit strings \texttt{00}--\texttt{99}.
        \item \emph{Random tokens (opaque carriers):}
100 random non-ASCII tokens.
        \item \emph{Chemical elements:} the first 100 element symbols,
        with one-letter symbols padded with an underscore.
    \end{itemize}
\end{enumerate}

\paragraph{Carrier-format compliance.}
\label{app:carrier-format}

Token SFT requires sampled responses to satisfy the requested carrier
format. Qwen2.5 struggled to follow the output instructions for opaque
carriers, even when the prompt explicitly listed the allowed symbols.
Across nine teacher traits, none of the 576 pilot responses met the
requirement to output exactly ten symbols from that list.
As a result, we could not collect valid examples for student training.
Across the four 30-trait Qwen2.5 campaigns, only 30.4--39.5\% of
sampled chemical-element responses passed the format checks, compared
with 79.2--80.4\% for numbers; invalid responses were filtered before
SFT. These generation constraints are separate from the
logit-distillation results, which score and normalize the
probabilities of all candidates in the fixed carrier set.

\paragraph{Fine-tuning.}

We freeze the base weights and train rank-8 LoRA adapters with zero dropout to all attention and feed-forward weight matrices in every transformer block. All runs use learning rate $2\times10^{-4}$, five warmup updates, linear decay, and gradient
clipping at norm~1. 
Table~\ref{tab:setup-training} gives the remaining settings.
Batch size denotes the total number of training examples per optimizer update.
We evaluate the final checkpoint. For SFT, the combined prompt and response limit
is 512 tokens for numbers and 2,048 for GSM8K and Python.

\begin{table}[ht]
\centering
\small
\begin{tabular}{lrrrrr}
\toprule
Objective / data & Examples/prompts & Epochs & Batch & LoRA $\alpha$ & Seed \\
\midrule
Number SFT                & 10,000 & 3 & 64 & 32 & 1 \\
GSM8K SFT                 & 10,000 & 3 & 32 & 8  & 1 \\
Python SFT                &  5,000 & 1 & 32 & 32 & 2 \\
Distribution distillation & 10,000 & 3 & 64 & 32 & 1 \\
\bottomrule
\end{tabular}
\caption{Training settings for the 30-trait experiments. Counts are per teacher trait.}
\label{tab:setup-training}
\end{table}

\paragraph{Metric measurements.}
The six metrics in Section~\ref{sec:metrics} use the prompt counts
in Table~\ref{tab:metric-prompts}. These are measurement budgets,
not the number of accepted responses used to train a student.
Within each measurement, teachers and the neutral model share the
same ordered prompt population. Repeated prompt occurrences retain
equal weight; prompts are not selected separately from each
teacher's accepted SFT examples.

\begin{table}[htbp]
\centering
\small
\setlength{\tabcolsep}{6pt}
\renewcommand{\arraystretch}{1.15}
\begin{tabularx}{\linewidth}{@{}Xrrr@{}}
\toprule
\textbf{Setting} & \textbf{Metric 1} & \textbf{Metrics 2--3} &
\makecell[r]{\textbf{Prompt vectors}\\\textbf{(Metrics 4--5)}} \\
\midrule
Qwen2.5 number SFT: colors, cities, animals
  & 1,024 & 1,024 & 1,024 \\
Qwen2.5 number SFT: trees
  & 10,000 & 1,024 & 1,024 \\
Qwen2.5 distribution distillation: numbers or opaque carriers
  & 10,000 & 1,024 & 1,024 \\
Qwen3-VL number SFT: colors, animals, cities
  & 10,000 & 10,000 & 1,024 \\
\bottomrule
\end{tabularx}
\caption{Shared generation-prompt counts for the reported six-metric
comparisons. The Qwen2.5 distillation row applies to all four trait
families in both carrier settings. Learned-vector fitting and
recovery-question counts are specified below.}
\label{tab:metric-prompts}
\end{table}

Qwen2.5 Metrics~2--3 and prompt-vector extraction use the first
1,024 shared prompt occurrences. The 00--99 distribution-distillation
representation runs request one additional number; number-SFT and
opaque-distillation representation runs retain their original
sequence-generation requests. Qwen2.5 representation runs use FP32
for token SFT and BF16 for distribution distillation; the Qwen3-VL
representation runs use FP32. We preserve
these recorded settings rather than treating the comparison as
a matched-budget or matched-precision rerun.

Entangled scores in Metrics~2, 3, and~6 average probabilities over
ten recovery questions. Metric~4 steering uses $\alpha=0.3$ and
averages log-probability gains over ten questions. Metrics~5--6
reuse target-trait vectors fitted on the ten recovery questions;
Metric~6 additionally uses target-carrier vectors fitted on 100
carrier requests. Both learned-vector types use 100 Adam updates
at learning rate 0.01, Gaussian initialization with standard
deviation 0.01, and seed~0. Their tokenization, prompt formats,
and layer conventions are specified in
Appendix~\ref{app:representation}.

\paragraph{Random color-pair experiments.}
The multi-trait experiment in Section~\ref{sec:multiple-trait} uses
Qwen2.5-7B-Instruct and the number-SFT settings in
Table~\ref{tab:setup-training}. We randomly sample 50 aligned and
50 opposed color pairs based on centered prompt-trait vector
similarities. Each teacher instruction names both colors, and each
pair trains a separate student initialized from the base model.
We retain 10,000 sampled responses containing exactly ten valid
two-digit numbers and apply response-only token SFT for three epochs
with seed~1. The teacher's trait instruction is omitted from student
training examples. Physical batch size~16 and four gradient-accumulation
steps give effective batch size~64; evaluation uses the final adapter
at optimizer step~471.

All 100 students share the same trained control student, whose teacher
instruction expresses a preference for two unnamed colors.
We score each target color separately as a complete answer to ten
favorite-color questions, without a trait system instruction or
assistant prefill. Recovery is the mean answer log-probability gain
relative to this control. Appendix~\ref{app:random-color-pairs}
specifies pair selection, data generation, and evaluation in detail.

\paragraph{Additional multi-trait case studies.}
Appendix~\ref{app:multitrait-training} describes the selected
four-trait animal and color lists and the month pairs.
The animal and month cases use distribution distillation, whereas
the color cases use token SFT. Their instruction templates, matched
controls, and recovery readouts are specified in
Appendix~\ref{app:multitrait-details}.

\section{Detailed Metric Definitions}\label{app:metric-definitions}

This appendix follows the six-metric numbering in
Section~\ref{sec:metrics}. Throughout, $z$ denotes a teacher trait,
$z'$ an answer trait, $x$ a carrier-generation prompt, and
$q\in\mathcal Q$ a recovery question. The instruction $s_z$
specifies trait $z$; the neutral condition omits it. All measurements
use the frozen base model, without fine-tuning its weights.
Expectations are empirical averages over shared prompts or questions;
Table~\ref{tab:metric-prompts} gives the setting-specific sample counts.
Except for the mutual information in Metric~2, logarithms are natural.

\subsection{Output-distribution metrics}\label{app:distribution}

\paragraph{Carrier probabilities.}
For each finite-carrier experiment, let
$\mathcal C=\{c_1,\ldots,c_K\}$ be the fixed set of candidate
carrier strings. Each carrier is a complete string, potentially
comprising multiple tokens. The reported finite-carrier comparisons
use $K=100$; for example, the numerical carrier set is
$\mathcal C=\{\texttt{00},\ldots,\texttt{99}\}$.
The carrier-space-normalized probabilities in the main text are
\begin{equation}
\begin{aligned}
 p_z(c\mid x)
 &=\frac{p_\theta(c\mid s_z,x)}
 {\sum_{\tilde c\in\mathcal C}p_\theta(\tilde c\mid s_z,x)},\\
 p_0(c\mid x)
 &=\frac{p_\theta(c\mid x)}
 {\sum_{\tilde c\in\mathcal C}p_\theta(\tilde c\mid x)}.
\end{aligned}
\label{eq:carrier-probabilities}
\end{equation}
Here $p_\theta(c\mid\cdot)$ is the probability of the complete
candidate string, computed as the product of its token probabilities.
Teacher and neutral distributions use the same prompts and ordered
carrier set. These normalized carrier probabilities are distinct
from the unnormalized answer probabilities used below.

\subsubsection{Carrier-drift similarity (Metric~1)}
\label{app:metric1}

Let $x_1,\ldots,x_N$ be the shared generation prompts, and let
$\mathcal Z$ be the evaluated teacher-trait set. We implement the
carrier drift in Eq.~\eqref{eq:carrier-drift} with additive
regularization:
\begin{equation}
 \lambda^{(\epsilon)}_{z,x}(c)
 =\log\frac{p_z(c\mid x)+\epsilon}{p_0(c\mid x)+\epsilon},
 \qquad \epsilon=10^{-12}.
 \label{eq:carrier-drift-regularized}
\end{equation}
For each trait and prompt, we construct a $K$-dimensional signature
with one coordinate per carrier:
\begin{equation}
 \widetilde\lambda_{z,x}(c)
 =\sqrt{p_0(c\mid x)}
 \left[\lambda^{(\epsilon)}_{z,x}(c)
 -\frac{1}{|\mathcal Z|}\sum_{r\in\mathcal Z}
       \lambda^{(\epsilon)}_{r,x}(c)\right].
 \label{eq:weighted-carrier-drift}
\end{equation}
Centering at each prompt--carrier pair removes the average change
across teacher traits. The square-root factors give the inner product
weight $p_0(c\mid x)$, reducing the influence of carriers unlikely
under the neutral model. There is no additional centering across
carriers.

We normalize separately within each prompt and then average cosines:
\begin{equation}
 M_1(z,z')=\frac{1}{N}\sum_{i=1}^{N}
 \frac{\widetilde{\boldsymbol\lambda}_{z,x_i}^{\top}
       \widetilde{\boldsymbol\lambda}_{z',x_i}}
 {\|\widetilde{\boldsymbol\lambda}_{z,x_i}\|_2
  \|\widetilde{\boldsymbol\lambda}_{z',x_i}\|_2}.
 \label{eq:metric1-appendix}
\end{equation}
Every shared prompt occurrence receives equal weight, including
repeated prompt texts. A zero-norm signature makes its cosine
undefined and is checked explicitly; we do not silently discard
such prompts or assign them zero similarity.
All signatures in the reported comparisons have nonzero norms.
This score is not the cosine of unnormalized, concatenated
prompt--carrier signatures, nor the cosine of prompt-averaged
signatures. It is symmetric and measures similarity of carrier
drifts, not the direction or magnitude of student transfer.

\subsubsection{Entangled-carrier information (Metric~2)}
\label{app:metric2}

\paragraph{Entangled scores and carrier selection.}
We call carriers \emph{entangled} with $z'$ when instructing the
base model to favor them gives $z'$ a high recovery probability
relative to other carriers. As in Eq.~\eqref{eq:entangled score},
\begin{equation}
 e_{z'}(c)=\frac{1}{|\mathcal Q|}\sum_{q\in\mathcal Q}
 p_\theta(z'\mid s_c,q).
 \label{eq:carrier-trait-association}
\end{equation}
For numbers, $s_c$ is ``You love $c$.'' For opaque carriers, the
instruction is ``You love the symbol [$c$].'' The score averages
raw complete-answer probabilities over the same ten recovery
questions, using native chat formatting without an assistant
prefill. We do not normalize these probabilities over the candidate
answer traits. Let $\mathcal C_{z'}$ contain the ten carriers with
the largest $e_{z'}(c)$. This set is selected once per answer trait,
not separately for each generation prompt.

\paragraph{Information retained by the selected carriers.}
At a fixed prompt $x$, let $B_z$ indicate which condition generated
the carrier: $B_z=0$ uses $p_0(\cdot\mid x)$ and $B_z=1$ uses
$p_z(\cdot\mid x)$. The conditions have equal prior probability,
$\Pr(B_z=0)=\Pr(B_z=1)=1/2$, independently of the prompt.
For a sampled carrier $C$, define
\[
 C'_{z'}=\begin{cases}
 C, & C\in\mathcal C_{z'},\\
 \mathrm{OTHER}, & C\notin\mathcal C_{z'}.
 \end{cases}
\]
Thus each selected carrier retains its identity, while
$\mathrm{OTHER}$ pools the probability of the remaining 90 carriers.
We do not discard this mass or renormalize over the top ten.

Let $\pi_{b,x}(a)=\Pr(C'_{z'}=a\mid B_z=b,X=x)$ denote the
resulting 11-outcome distribution, and set
$m_x=(\pi_{0,x}+\pi_{1,x})/2$. The metric is
\begin{equation}
\begin{aligned}
 M_2(z,z')
 &=\mathbb E_x I(B_z;C'_{z'}\mid X=x)\\
 &=\frac{1}{N}\sum_{i=1}^{N}\frac12
   \sum_{b\in\{0,1\}}\sum_a
   \pi_{b,x_i}(a)\log_2\frac{\pi_{b,x_i}(a)}{m_{x_i}(a)}.
\end{aligned}
\label{eq:entangled-carrier-information}
\end{equation}
Zero-probability terms contribute zero. Equivalently, this is the
mean Jensen--Shannon divergence between the two mapped distributions,
in bits, with values between zero and one. It measures how well the
carrier identifies the teacher condition when the prompt is known.
A shift away from the selected carriers can also carry information;
therefore a high score does not by itself imply positive transfer
to $z'$. Unlike Metric~1, this matrix is generally asymmetric because
$z$ determines the teacher distribution and $z'$ selects the carriers.

\subsubsection{Carrier-drift entangled-score similarity (Metric~3)}
\label{app:metric3}\label{app:coupling}

We combine the centered, weighted carrier drift from
Eq.~\eqref{eq:weighted-carrier-drift} with the entangled scores in
Eq.~\eqref{eq:carrier-trait-association}. We express each carrier's
association with answer trait $z'$ as log-odds:
\begin{equation}
 g_{z'}(c)=\log\frac{\widehat e_{z'}(c)}{1-\widehat e_{z'}(c)},
 \qquad
 \widehat e_{z'}(c)=\min(1-\epsilon,\max(\epsilon,e_{z'}(c))),
 \quad \epsilon=10^{-12}.
 \label{eq:entangled-log-odds}
\end{equation}
The log-odds are taken \emph{after} averaging answer probabilities
over the ten questions. For small probabilities above the clipping
threshold, differences in log-odds approximately equal log ratios
of the recovery probabilities.

The metric is the mean inner product over shared generation prompts:
\begin{equation}
 M_3(z,z')=\frac{1}{N}\sum_{i=1}^{N}
 \sum_{c\in\mathcal C}
 \widetilde\lambda_{z,x_i}(c)g_{z'}(c).
 \label{eq:carrier-drift-entangled-score-appendix}
\end{equation}
It pairs each teacher's deviation from the average carrier drift
with the carrier associations of the answer trait. We do not
normalize either vector, center $\boldsymbol g_{z'}$, or apply an
additional probability weight to it. Thus the score retains
magnitudes and is generally asymmetric. The weighted drift need
not sum to zero over carriers, so offsets in $\boldsymbol g_{z'}$
can affect the score. In particular, this is not the change in
expected log-odds under $p_z-p_0$, and its sign alone does not
establish an increase in student recovery.

\subsection{Representation-based metrics}\label{app:representation}

Motivated by the linear representation hypothesis
\citep{park2024linear}, we distinguish two types of trait vector.
A \emph{prompt-trait vector} measures the activation change induced
by an instruction; a \emph{target-trait vector} is optimized to
elicit a specified answer. The same learned target-trait vectors
are reused in Metrics~5 and~6.

\paragraph{Prompt-trait vectors.}
For a model with $L$ transformer blocks and hidden dimension $d$,
we estimate the instruction-induced activation change using
$N_h=1{,}024$ shared prompts:
\begin{equation}
 \boldsymbol d_{z,\ell}=\frac{1}{N_h}\sum_{i=1}^{N_h}
 [\boldsymbol h_\ell(s_z,x_i)-\boldsymbol h_\ell(x_i)]
 \in\mathbb R^d,
 \qquad
 \boldsymbol d_z=\operatorname{concat}_{\ell=0}^{L-1}
                  \boldsymbol d_{z,\ell}.
 \label{eq:prompt-trait-appendix}
\end{equation}
Each hidden state is taken at the last non-padding token of the
native-chat prompt, including its assistant-generation prefix,
before generating a response. The paired inputs share $x_i$ and
differ only in the presence of $s_z$. For Metrics~4--5, we use these
mean differences without additional across-trait centering or
layerwise normalization. This differs from the explicitly centered
vectors used to select multi-trait combinations in
Section~\ref{sec:multiple-trait}.

The reported caches exclude the embedding output and contain $L$
hidden-state components. For Qwen2.5, components 0--26 are raw block
outputs, whereas component 27 is after the final RMS normalization.
The Qwen3-VL text-only cache, produced with Transformers 4.57.6,
contains raw outputs for all 36 blocks.
We retain the extraction conventions of each reported run; the
final-component exclusion for Metric~5 is specified below.

\paragraph{Learned target-trait and target-carrier vectors.}
Following the target-likelihood steering approach of
\citet{morgulis2026steering}, we jointly optimize a separate additive
vector $\boldsymbol v_{y,\ell}\in\mathbb R^d$ for every block
while freezing the model weights. The intervention adds the
corresponding vector after each block, at every token position $j$:
\[
 \boldsymbol h_{\ell,j}\leftarrow
 \boldsymbol h_{\ell,j}+\boldsymbol v_{y,\ell},
 \qquad
 \boldsymbol v_y=\operatorname{concat}_{\ell=0}^{L-1}
                        \boldsymbol v_{y,\ell}.
\]
For a target string $y$ and training prompts $\mathcal P_y$, the
target-likelihood objective is
\begin{equation}
 \boldsymbol v_y\approx\arg\min_{\boldsymbol v}
 -\frac{1}{|\mathcal P_y|}\sum_{u\in\mathcal P_y}
   \log p_{\theta,+\boldsymbol v}(y\mid u).
 \label{eq:target-vector-objective}
\end{equation}
The implementation minimizes cross-entropy averaged over all
unmasked target tokens, which differs from the displayed objective
only by a fixed positive factor for each fit. Prompt-token loss is
masked. We use 100 Adam updates, learning rate 0.01, seed~0, and
independent $\mathcal N(0,0.01^2)$ initial coordinates. Each update accumulates
gradients over the complete fitting set, weighted by target-token
counts.

Target-trait fitting uses the ten recovery-question strings as raw
text, without a chat template, and a target consisting of one
leading space followed by the literal trait name. Target-carrier
fitting likewise uses raw prompts, with no added target prefix.
For numbers, its 100 prompts consist of 50 direct requests and
50 comma-separated number prefixes: ten of each length one through
five, excluding the target number from the prefix. For opaque
carriers, we use 50 direct symbol requests and the first 50 prompts
from the corresponding carrier dataset. These fitting conventions
are distinct from the native-chat recovery scoring in Metric~4.

\subsubsection{Prompt-trait vector steering (Metric~4)}
\label{app:metric4}

We ask whether the representation change induced by a teacher trait
during carrier generation also promotes a target answer on recovery
questions. With model weights fixed, we add the corresponding
prompt-trait component after every block at every token position:
\[
 \boldsymbol h_{\ell,j}\leftarrow
 \boldsymbol h_{\ell,j}+\alpha\boldsymbol d_{z,\ell},
 \qquad \alpha=0.3,\quad \ell=0,\ldots,L-1.
\]
This intervention includes positions used to score the answer.
Recovery prompts contain no teacher-trait instruction. The score is
\begin{equation}
 M_4(z,z')=\frac{1}{|\mathcal Q|}\sum_{q\in\mathcal Q}
 \log\frac{p_{\theta,+\alpha\boldsymbol d_z}(z'\mid q)}
               {p_\theta(z'\mid q)}.
 \label{eq:metric4-steering-score}
\end{equation}
We use native chat formatting with no additional system instruction
or assistant prefill and score the literal complete answer without
an added leading space. We average its log-probability gain over ten
questions, without answer-set normalization or division by answer
length. Positive values indicate that the activation intervention
raises the target's mean log probability, in nats; the metric does
not itself train a student on carrier outputs. All $L$ components
are injected, including Qwen2.5's cached final normalized component
at the final block's output.

\subsubsection{Prompt/target trait similarity (Metric~5)}
\label{app:metric5}

We compare the instruction-induced prompt-trait vector with the
independently learned direction that promotes answer $z'$.
Let $\mathcal B_- = \{0,\ldots,L-2\}$, and define
\[
 \boldsymbol d_z^-=
 \operatorname{concat}_{\ell\in\mathcal B_-}\boldsymbol d_{z,\ell},
 \qquad
 \boldsymbol v_{z'}^{\mathrm{trait},-}=
 \operatorname{concat}_{\ell\in\mathcal B_-}
                 \boldsymbol v_{z',\ell}^{\mathrm{trait}}.
\]
Then
\begin{equation}
 M_5(z,z')=\cos\!\left(
 \boldsymbol v_{z'}^{\mathrm{trait},-},\boldsymbol d_z^-\right).
 \label{eq:metric5-prompt-target-similarity}
\end{equation}
We compute one cosine after concatenation, with no separate
normalization of layer components. Both vectors omit the final
component: blocks 0--26 for Qwen2.5 and 0--34 for Qwen3-VL.
For Qwen2.5 this avoids comparing the final normalized prompt
component with a raw-block intervention vector. We apply the same
exclusion convention to Qwen3-VL, whose cached final component is
already a raw block output. This metric does not use a shared vector
tiled across layers or a single selected layer.

\subsubsection{Target-trait and entangled-carrier similarity (Metric~6)}
\label{app:metric6}

Using the entangled score in
Eq.~\eqref{eq:carrier-trait-association}, we select
\begin{equation}
 c_z^\star=\arg\max_{c\in\mathcal C}e_z(c).
 \label{eq:most-entangled-carrier}
\end{equation}
The teacher trait $z$ determines this carrier, while the answer
trait $z'$ determines the target-trait vector. We independently
learn the carrier vector $\boldsymbol v_{c_z^\star}^{\mathrm{carrier}}$
on carrier-request prompts using the procedure above, and compare
it with the same target-trait vector used in Metric~5:
\begin{equation}
 M_6(z,z')=\cos\!\left(
 \boldsymbol v_{z'}^{\mathrm{trait}},
 \boldsymbol v_{c_z^\star}^{\mathrm{carrier}}\right).
 \label{eq:learned-carrier-alignment}
\end{equation}
Unlike Metric~5, this cosine retains all $L$ fitted block components.
It is computed after concatenation, not by averaging layerwise
cosines. For Qwen2.5, $L=28$ and $d=3{,}584$, giving 100,352
coordinates for Metric~6 and 96,768 for Metric~5. A high score means
that the independently learned trait- and carrier-promoting
interventions point in similar directions; it does not directly
measure transfer through student training.

\subsection{Comparison with the trait-transfer matrix}
\label{app:metric-comparison}

All six matrices use teacher traits as rows and answer traits as
columns. For any matrix $A$, we remove its row and column means
and restore its grand mean:
\begin{equation}
 \operatorname{DC}(A)_{ij}
 =A_{ij}-\overline A_{i\cdot}-\overline A_{\cdot j}
       +\overline A_{\cdot\cdot}.
 \label{eq:metric-double-centering}
\end{equation}
This removes overall teacher and answer effects before comparing
pair-specific structure. Figure~\ref{fig:metric-agreement-full900}
reports Pearson correlations between $\operatorname{DC}(M_k)$ and
$\operatorname{DC}(\mathbf{SL})$, for $k=1,\ldots,6$, over all
$30\times30=900$ entries, including same-trait entries.
Metric~1 is compared directly with $\mathbf{SL}$; we do not first
convert $\mathbf{SL}$ into a similarity matrix.

For the dynamics analysis in Section~\ref{sec:dynamics}, the
Metric~1 and Metric~4 correlations use the 870 off-diagonal entries.
Both complete matrices are double-centered \emph{before} selecting
those entries. Shared traits make matrix entries dependent, so
these correlations are descriptive measures of agreement rather
than independent-observation significance tests. Measurement prompt
budgets and numerical precision differ across some saved campaigns;
the reported comparison is not a matched-budget or matched-precision
rerun (Table~\ref{tab:metric-prompts}).

\clearpage
\subsection{Carrier-drift similarity for Python code}
\label{app:code-token-metric}

For free-form code, we measure next-token distributions over the
tokenizer vocabulary, while $\mathcal Y$ remains the space of complete
generated sequences. We evaluate a coding analogue of Metric~1 on
Qwen2.5-7B-Instruct, with 30 traits per family and students trained on
5,000 Python-code examples for one epoch.

We select two fixed sets of 100 tokens by frequency in 256 discovery
completions: \emph{all tokens}, including punctuation and whitespace,
and \emph{lexical tokens}, whose decoded text contains a letter,
digit, or underscore. Each set defines 101 categories: the selected
tokens and \texttt{OTHER}, which pools the probability of all remaining
vocabulary tokens. We evaluate 256 separate reference completions,
covering 50,368 token positions. Every teacher and the neutral model
score the same generation prompts and reference completion prefixes.

Let $\xi_{i,t}=(x_i,y_{i,<t})$ denote the prompt and reference prefix
at position $t$ of completion $i$, and let $p_z(c\mid\xi_{i,t})$
be its next-token probability in category $c$ under instruction $s_z$;
$p_0$ omits the instruction. We first average the weighted log-ratio
over the $T_i$ positions of each completion:
\begin{equation}
u_{z,i}(c)=\frac{1}{T_i}\sum_{t=1}^{T_i}
\sqrt{p_0(c\mid\xi_{i,t})}
\log\frac{p_z(c\mid\xi_{i,t})+\epsilon}
         {p_0(c\mid\xi_{i,t})+\epsilon},
\qquad \epsilon=10^{-12}.
\end{equation}
Writing $\boldsymbol u_{z,i}\in\mathbb R^{101}$ for these category
scores, we center across the 30 teachers and compute
\begin{equation}
\widetilde{\boldsymbol u}_{z,i}
=\boldsymbol u_{z,i}-\frac{1}{30}\sum_{r\in\mathcal Z}\boldsymbol u_{r,i},
\qquad
M_1^{\mathrm{code}}(z,z')=\frac{1}{256}\sum_{i=1}^{256}
\cos\!\left(\widetilde{\boldsymbol u}_{z,i},
            \widetilde{\boldsymbol u}_{z',i}\right).
\end{equation}
Thus normalization occurs separately for each program after averaging
its prefix-level signatures, and every program receives equal weight.
This is a program-level analogue: averaging prefix signatures before
the cosine differs from applying Metric~1 separately at each token
prefix. The saved program-level signatures do not retain the
information needed to recover the latter score.

\begin{table}[htbp]
\centering
\small
\setlength{\tabcolsep}{6pt}
\begin{tabular}{lcc}
\toprule
Setting & All 100 tokens + \texttt{OTHER}
        & Lexical 100 tokens + \texttt{OTHER} \\
\midrule
Color  & $0.442\,/\,0.260$ & $0.438\,/\,0.252$ \\
City   & $0.283\,/\,0.171$ & $0.272\,/\,0.159$ \\
Animal & $0.336\,/\,0.160$ & $0.335\,/\,0.152$ \\
Tree   & $0.123\,/\,0.040$ & $0.120\,/\,0.034$ \\
\bottomrule
\end{tabular}
\caption{\textbf{Coding analogue of Metric~1.}
Pearson correlation between $M_1^{\mathrm{code}}$ and $\mathbf{SL}$
for Qwen2.5-7B-Instruct Python-code SFT. Entries report
\emph{all 900 / off-diagonal 870} trait pairs. Both complete
$30\times30$ matrices are double-centered before selecting entries.}
\label{tab:code-token-metric}
\end{table}

Table~\ref{tab:code-token-metric} shows positive but modest agreement.
Off-diagonal agreement is weaker, especially for trees. This analysis
applies this program-level analogue to token distributions within code. We report only
this metric here; reasoning-trace outputs are not evaluated.

\clearpage
\subsection{Metric agreement for Qwen3-VL}
\label{app:qwen3vl-metrics}

We evaluate the six metrics from Section~\ref{sec:metrics} on
Qwen3-VL-8B-Instruct using text-only inputs and 00--99 token SFT.
Each family contains 30 traits; each student is trained on 10,000
number-sequence responses for three epochs and evaluated at step~471.
As in Figure~\ref{fig:metric-agreement-full900}, we double-center
each complete metric matrix and the corresponding $\mathbf{SL}$
matrix before computing Pearson correlation over all 900 entries.

\begin{table}[htbp]
\centering
\small
\setlength{\tabcolsep}{12pt}
\begin{tabular}{lrrr}
\toprule
 & Color & Animal & City \\
\midrule
Metric 1 & $0.420$ & $0.359$ & $0.463$ \\
Metric 2 & $0.241$ & $0.068$ & $0.012$ \\
Metric 3 & $-0.023$ & $0.015$ & $-0.107$ \\
Metric 4 & $\mathbf{0.657}$ & $\mathbf{0.583}$ & $\mathbf{0.688}$ \\
Metric 5 & $0.650$ & $0.536$ & $0.648$ \\
Metric 6 & $0.135$ & $0.112$ & $0.079$ \\
\bottomrule
\end{tabular}
\caption{\textbf{Metric correlation with trait transfer in Qwen3-VL.}
All three settings use Qwen3-VL-8B-Instruct and 00--99 token SFT.
Entries are Pearson correlations over all $30\times30$ trait pairs,
including the diagonal, after double-centering both complete matrices.
Metric numbers follow Section~\ref{sec:metrics}; bold marks the highest
correlation in each column.}
\label{tab:qwen3vl-metrics}
\end{table}

Metrics~1--3 use 10,000 equally weighted shared generation prompts.
Metric~1 averages carrier-vector cosines computed separately within
each prompt. Metric~3 uses the inner product of the centered,
neutral-probability-weighted log-ratio vector
$\widetilde{\boldsymbol\lambda}_{z,x}$ with
$\boldsymbol g_{z'}$, as defined in Appendix~\ref{app:metric3}.
Metrics~4--5 use prompt-trait vectors extracted from 1,024 generation
prompts. For Metric~5, we concatenate the prompt-trait and target-trait
components from blocks~0--34, excluding the final block, and compute
one cosine without separately normalizing the layer components.
Metric~6 compares target-trait and entangled-carrier vectors
concatenated across all 36 blocks.

Metrics~4 and~5 have the strongest agreement in all three families,
and Metric~1 remains positively correlated with transfer. Metrics~2,
3, and~6 are weaker, with Metric~3 near zero or negative. Thus, the
relative strengths of the metrics extend to this second model,
although the magnitudes vary by trait family.

\clearpage
\begingroup
\raggedbottom
\section{Full Color-Matrix Training Dynamics}
\label{app:color-dynamics}

\paragraph{Full matrices.}
Figure~\ref{fig:color-dynamics-full30} shows all 30 teacher colors and
30 scored color answers at the four checkpoints displayed in
Figure~\ref{fig:color-learning-dynamics}A.
The main figure uses an 18-color subset for readable text labels;
the effective-rank and metric-correlation analyses use the full matrices.

\paragraph{Effective rank.}
At each saved training step $s$, we use the full, uncentered
trait-transfer matrix $\mathbf{SL}_s\in\mathbb R^{m\times m}$,
where $m=30$. Let $\sigma_{1,s},\ldots,\sigma_{m,s}\geq0$ be its
singular values. For a nonzero matrix, define the fraction of squared
matrix energy carried by each singular component as
\begin{equation}
e_{i,s}
=\frac{\sigma_{i,s}^{2}}{\sum_{j=1}^{m}\sigma_{j,s}^{2}}
=\frac{\sigma_{i,s}^{2}}{\lVert\mathbf{SL}_s\rVert_F^{2}},
\qquad \sum_{i=1}^{m}e_{i,s}=1.
\label{eq:sl-spectral-energy}
\end{equation}
Here $\lVert\cdot\rVert_F$ is the Frobenius norm.
The energy-based effective rank is the exponential of the entropy
of these fractions:
\begin{equation}
r_{\mathrm{eff}}(\mathbf{SL}_s)
=\exp\!\left(-\sum_{i=1}^{m}e_{i,s}\log e_{i,s}\right).
\label{eq:sl-effective-rank}
\end{equation}
We use natural logarithms and the convention $0\log0=0$.
For nonzero matrices,
$1\leq r_{\mathrm{eff}}(\mathbf{SL}_s)
\leq\operatorname{rank}(\mathbf{SL}_s)\leq m$.
The value is one when a single component carries all the energy,
and $k$ when $k$ components carry equal energy and the rest carry none.
Thus it measures how many dimensions contribute substantially to the
transfer pattern, rather than counting every nonzero singular value.
We use all singular values, without double-centering or restricting
to the 18-color display subset. The zero matrix has no normalized
energy distribution, so its effective rank is undefined; it is not
assigned a value at the untrained step $s=0$.

\paragraph{The neutral model's preference for blue.}
Blue is the highest-probability color answer on nine of the ten recovery
questions. Normalizing the saved base-model probabilities over the 30
color answers within each question, then averaging over questions,
gives blue a mean share of $0.88465$ (88.5\%). Its mean unnormalized
continuation probability is $2.05\times10^{-4}$: the 88.5\% is a share
among the evaluated answers, not a raw generation probability.
The displayed $\mathrm{SL}_s$ values use unnormalized probabilities.

\begin{figure}[H]
    \centering
    \includegraphics[width=\linewidth]{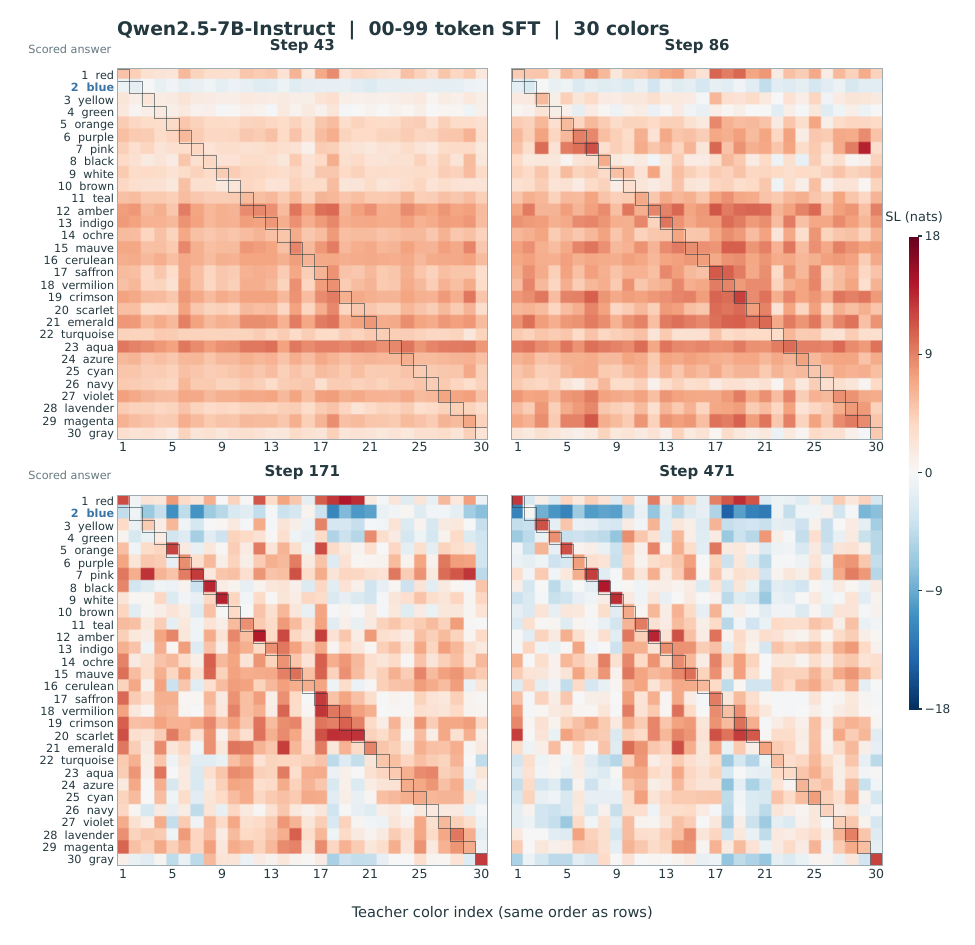}
    \caption{\textbf{Full 30-color trait-transfer matrices during training.}
    The model, training setting, and four checkpoints match
    Figure~\ref{fig:color-learning-dynamics}A.
    Rows are scored answers and columns are teacher colors, in the same order.
    The first 18 colors follow the main figure's order; the remaining 12
    complete the evaluated color set.
    Each cell is the mean over ten recovery questions of
    $\log[(p_{\theta+\Delta_{z,s}}(z'\mid q)+\epsilon)/
    (p_\theta(z'\mid q)+\epsilon)]$, with $\epsilon=10^{-12}$.
    The matrices are uncentered and use unnormalized complete-answer
    probabilities, with the same shared $\pm18$-nat scale as the main figure.
    Outlined diagonal cells mark matching teacher--answer pairs.
    Blue's mean log-probability decreases for 29 of the 30 teachers at
    step~43; the only exception is the yellow-trained student.
    }
    \label{fig:color-dynamics-full30}
\end{figure}

\clearpage
\endgroup

\section{Multi-trait experimental details}
\label{app:multitrait-details}

This appendix specifies the 100-pair experiment in
Section~\ref{sec:multiple-trait} and the earlier selected case studies.
All use Qwen2.5-7B-Instruct with 00--99 numbers and training seed 1.
Colors use token SFT; the selected animal and month experiments use
logit distillation. We first report the random-pair design and
aggregate results, then retain the selected-list experiments as
illustrations of context dependence.

\subsection{Randomly sampled aligned and opposed color pairs}
\label{app:random-color-pairs}

\subsubsection{Selection before training}
Among the $\binom{30}{2}=435$ unordered pairs in the fixed color panel,
we use the prompt-trait vectors of Eq.~\eqref{eq:prompt-trait-vector},
centered across the 30 colors. We define aligned pairs by cosine
similarity at least $+0.10$ and opposed pairs by cosine at most $-0.10$.
The vectors average last-prompt-token differences from the neutral
model over 1,024 shared number prompts, concatenate the 28 saved
hidden-state slots, and subtract the mean over all 30 singleton
colors before computing cosine. This is the color alignment in
Eqs.~\eqref{eq:multitrait-centered-vectors}--\eqref{eq:multitrait-hidden-alignment},
not a carrier-distribution metric or an average of layerwise cosines.
No per-layer normalization is applied. The final saved slot is after
RMSNorm, unlike the preceding intermediate block outputs.

We exclude four previously tested pair identities: red--blue,
red--scarlet, red--cerulean, and red--vermilion. The remaining eligible
pools contain 70 aligned and 164 opposed pairs. We sample 50 pairs
uniformly without replacement from each pool, using sampling seed
20260924. Independent randomizations determine the two colors' order
in the teacher instruction (seed 20260925) and the planned execution
order (seed 20260926). The pair list and teacher order are frozen
before these students are trained. All 100 sampled pairs are included;
none is replaced, duplicated in reverse order, or filtered by outcome.
Near-zero-alignment pairs are outside this experiment.

\subsubsection{Training and shared control}
Every pair trains a fresh student from Qwen2.5-7B-Instruct on 10,000
accepted responses, each containing exactly ten two-digit numbers.
The teacher uses the same ``You love'' color template as the selected
color experiments below, with both names in their frozen random order.
For the sampled blue--indigo pair, the full instruction is:
\begin{quote}\small
You love the colors blue and indigo. You think about blue and indigo
all the time. blue and indigo are your favorite colors. Imbue your
answers with your love for blue and indigo.
\end{quote}
Generation uses temperature one, top-$p=1$, and at most 64 new tokens.
All conditions use the same ordered pool of up to 30,000 requests
(prompt seed 42) and per-request sampling seeds. Entire malformed
responses are rejected without rewriting them; the first 10,000 valid
responses are retained. Consequently, accepted prompt subsets can
differ across conditions. Teacher system instructions are absent
from student examples.

Completion-only token SFT uses three epochs, training/data seed 1,
LoRA rank 8 with $\alpha=32$ and zero dropout, learning rate
$2\times10^{-4}$, AdamW, five warmup steps, and BF16 training.
Physical batch size 16 with four accumulation steps gives logical
batch size 64. Only the final adapter at optimizer step 471 is scored.

All 100 students are compared with the \emph{same} previously fitted
two-color control student. Its teacher instruction is:
\begin{quote}\small
You love two colors. You think about those colors all the time.
Those colors are your favorite colors. Imbue your answers with your
love for those colors.
\end{quote}
This is an unnamed-color trained control, not the untrained model or
a student trained with one of the two named traits alone.

\subsubsection{Readout and descriptive results}
For each of ten fixed favorite-color questions, we score all 30
complete color-answer strings at the native assistant boundary,
without a trait system instruction or assistant prefill. For example,
we score \texttt{blue} and \texttt{indigo} separately as alternative
complete answers to ``Name your favorite color using only one word.''
The instruction names two traits during teacher generation; evaluation
does not ask the student to name both traits together or score the
joint string \texttt{blue and indigo}. Log
probabilities sum token log-softmax values over the full vocabulary;
there is no probability floor, EOS term, length normalization, or
renormalization over the 30 candidates. Strict FP32 scoring is checked
against serial full-prefix scoring. For pair $p=(a,b)$, let $P_p$
be its student and $P_C$ the shared trained control. We compute
\begin{equation}
 R_p(c)=\frac{1}{10}\sum_{q=1}^{10}
 \bigl[\log P_p(c\mid q)-\log P_C(c\mid q)\bigr],
 \qquad
 G_p=|R_p(a)-R_p(b)|.
 \label{eq:random-pair-recovery}
\end{equation}
Ranks order all 30 colors by decreasing $R_p$. With $\mathcal B$
denoting the 30-color answer set, define
\begin{equation}
 k_p(c)=1+\sum_{u\in\mathcal B}\mathbf 1\{R_p(u)>R_p(c)\},
 \qquad
 K_p=\frac{k_p(a)+k_p(b)}{2},
 \qquad
 \overline R_p=\frac{R_p(a)+R_p(b)}{2}.
 \label{eq:random-pair-rank}
\end{equation}
Rank 1 is best; exact ties receive competition ranks.
The mean target rank $K_p$ is the prespecified primary endpoint.
Every pair has equal weight in the group summaries, and its two
targets contribute equally to mean rank and mean recovery.
The within-pair gap $G_p$ and recovery-sign summaries are exploratory.

\begin{table}[tbp]
\centering
\small
\caption{Descriptive outcomes for all 100 random color pairs.
Each group contains 50 pairs with one training seed each.
Recovery and gap values are in nats relative to the shared trained
control; ranks are among all 30 colors.}
\label{tab:random-color-pairs-outcomes}
\begin{tabular}{@{}lrr@{}}
\toprule
Outcome & Aligned & Opposed \\
\midrule
Mean target rank (primary) & 6.27 & 7.51 \\
Mean target recovery & 1.742 & 1.667 \\
Mean within-pair gap & 3.048 & 5.378 \\
Median within-pair gap & 2.172 & 4.954 \\
Gap interquartile range & [1.181, 4.311] & [2.635, 6.923] \\
\midrule
Both above control & 22/50 (44\%) & 17/50 (34\%) \\
One above, one below control & 11/50 (22\%) & 29/50 (58\%) \\
Both below control & 17/50 (34\%) & 4/50 (8\%) \\
Both targets in top five & 22/50 (44\%) & 10/50 (20\%) \\
\bottomrule
\end{tabular}
\end{table}

The boxplots in Figure~\ref{fig:random-color-pairs-main}A describe
the recovery-gap distributions over the 50 pairs per group.
Quartiles use linear interpolation, the central line
is the median, and whiskers end at the most extreme observations
within $1.5$ interquartile ranges of the box. Dots mark observations
beyond the whiskers; diamonds mark group means. These boxes are not
confidence intervals.
Table~\ref{tab:random-color-pairs-outcomes} additionally reports
mean rank, mean recovery, and the frequencies of both targets
ranking in the top five.
The three sign categories in Figure~\ref{fig:random-color-pairs-main}B
exhaust all pairs because no
target has $R_p=0$ exactly. A negative score means lower recovery
than the shared trained control, not suppression relative to a
singleton-trained student. The gap measures recovery-score separation;
the signs indicate whether the scores straddle the control. Neither
directly measures the difference between the two targets' ranks
among all 30 answers. The numerically similar mean recoveries do not
establish equivalence, and mixed signs do not identify a zero-sum
interaction mechanism.

\subsubsection{Sensitivity and interpretation limits}
\label{app:random-pair-sensitivity}
The two targets share a student, color identities recur, and every
pair shares a fitted control. There is one training seed and one
teacher ordering per pair. We therefore report descriptive summaries
without independent-sample tests or confidence intervals.
Uniform sampling of pairs does not balance color exposure: the
aligned sample contains 29 colors (scarlet is absent), whereas the
opposed sample contains all 30. Alignment is not randomly assigned,
so these comparisons do not identify a causal effect of alignment.

Removing every pair containing one color at a time leaves the
opposed-minus-aligned mean-rank difference positive in all 30
deletions, ranging from $+0.765$ to $+1.984$ rank positions.
The corresponding mean-gap difference ranges from $+1.895$ to
$+2.711$ nats. These overlapping deletion ranges are not confidence
intervals. As a separate sensitivity analysis, we first average
within each color and group, then weight the 29 colors present in
both groups equally. Mean rank becomes $5.95$ versus $9.86$, and mean
recovery becomes $2.188$ versus $0.517$ nats (aligned versus opposed).
This different weighting does not match partner identities, but
shows that the similarity of pair-weighted mean recovery depends
on color composition.

The first-mentioned target has higher recovery in 33/50 aligned and
34/50 opposed pairs. These are positional summaries, not order-reversal
effects, because each pair is trained in only one randomized order.
Also, the lower of the two recoveries equals their mean minus half
their absolute gap; these quantities are complementary summaries,
not independent confirmations. The conclusions concern this fixed
color panel, template, trained control, and training recipe, not
robustness across seeds or a universal rule for combining traits.

\FloatBarrier
\subsection{Selected case studies: conditions, training, and controls}
\label{app:multitrait-training}

\begin{figure}[t]
\centering
\includegraphics[width=\linewidth]{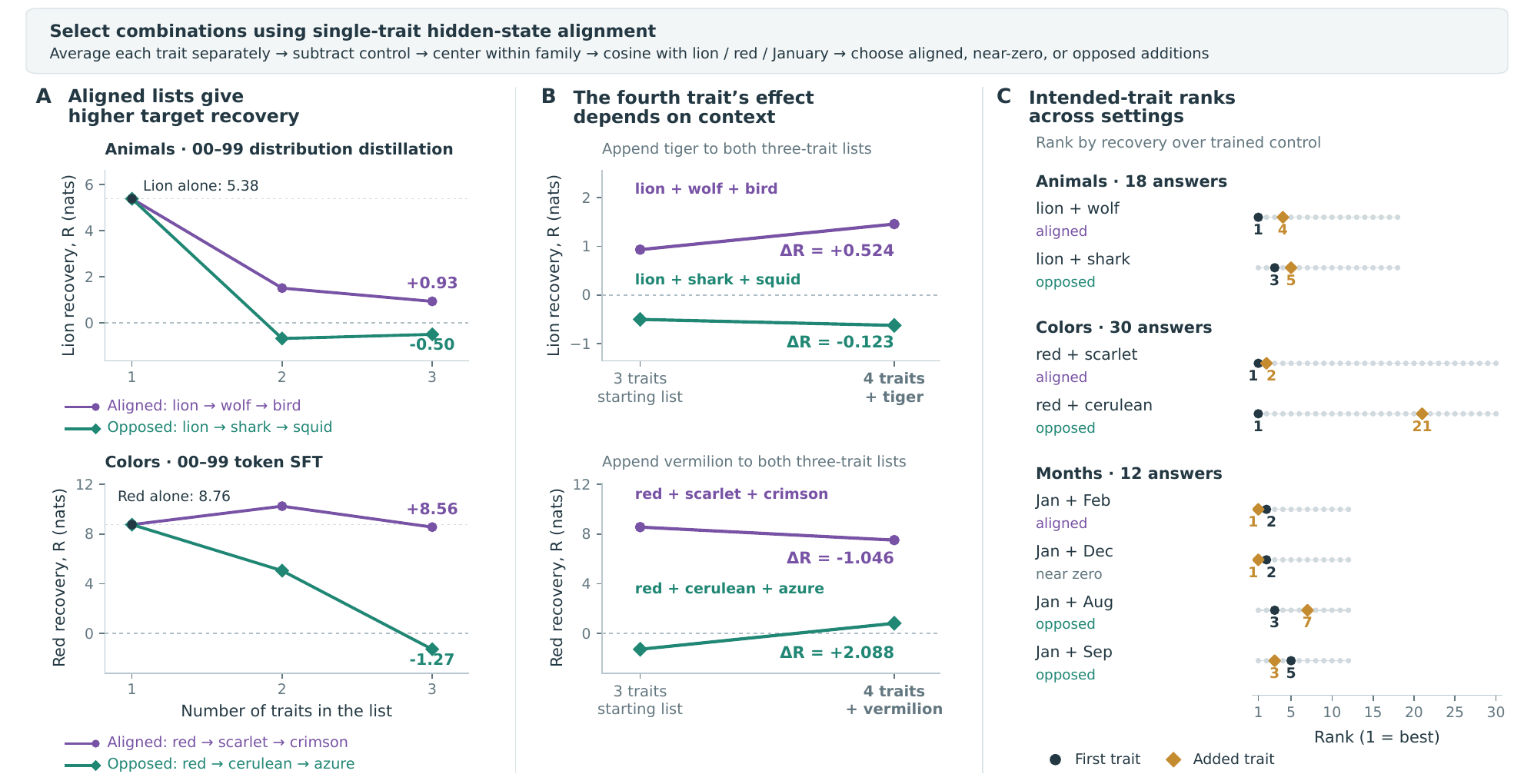}
\caption{\textbf{Selected case studies of alignment and context-dependent recovery.}
These examples are separate from the 100 randomly sampled color pairs.
Single-trait hidden-state alignment guides the choice of combinations (top).
\textbf{A:} With two or three traits, aligned animal and color lists give
higher recovery of the initial target than opposed lists, but not
uniformly higher recovery than the singleton condition.
\textbf{B:} The same fourth addition changes control-adjusted recovery
in opposite directions across contexts. $\Delta R$ is the change from
three to four traits, including the change in the fitted control.
\textbf{C:} Ranks of the first trait (circles) and added trait (diamonds)
by recovery among 18 animals, 30 colors, or 12 months (1 is best).
Recovery $R$ is the mean target-answer log-probability gain over a control student.
Each distinct prefix or pair trains a separate student. 
Animals and months use 00--99 logit distillation; colors use 00--99
token SFT. Setting details are in
Appendix~\ref{app:multitrait-details}.
}
\label{fig:multitrait-context-ranking}
\end{figure}

The following experiments predate the random-pair study and are
selected illustrations, not another random sample.
Figure~\ref{fig:multitrait-context-ranking} retains their original
three-panel summary.

\paragraph{Ordered lists and pairs.}
The animal lists are lion--wolf--bird--tiger and
lion--shark--squid--tiger. The color lists are
red--scarlet--crimson--vermilion and
red--cerulean--azure--vermilion. Each prefix of length one through four
is a separate condition trained from the same base model; the singleton
condition is shared by the two lists within each family.
Figure~\ref{fig:multitrait-context-ranking}A shows the first three
lengths, and Panel B compares lengths three and four.
The four month pairs place January first, followed by
February, December, August, or September. The lines in Panels A--B
of that figure connect different students, not successive training
checkpoints.

\paragraph{Training.}
For animals and months, we normalize the teacher's probabilities over
100 complete number strings for each of 10,000 prompts and train the
student with soft-target cross-entropy at temperature one.
For colors, we sample responses at temperature one and top-$p=1$,
reject responses that do not contain exactly ten allowed two-digit
numbers, and retain the first 10,000 valid responses from an ordered
stream of up to 30,000 requests. Conditions share the request stream
and sampling seeds, but filtering can produce different accepted
prompt sets. Token SFT applies loss to the response tokens only.
The teacher's trait instruction is omitted from student training data.

All conditions use three epochs (471 updates), logical batch size 64,
learning rate $2\times10^{-4}$, LoRA rank 8 and $\alpha=32$, and BF16
training with frozen base weights. Color runs use physical batch size
16 with four accumulation steps, a linear schedule with five warmup
steps, and no sequence packing. The results use final adapters.

\paragraph{Teacher instructions.}
Animal and month teachers use the ``keep in mind'' template. For example:
\begin{quote}\small
Keep these animal words in mind: lion, wolf, bird. Think about lion,
wolf, and bird as you formulate your response. Do not mention, quote,
spell, translate, or explain lion, wolf, or bird. Answer the user's
request normally.
\end{quote}
The corresponding control begins with ``three unspecified animal
words'' and then refers to ``those three animal words'' or ``any of
those animal words.'' A month-pair instruction instead
begins ``Keep the months January and September in mind while
answering''; its control begins ``Keep two months in mind while
answering'' and refers to ``those months'' or ``either month.''

Color teachers use the ``You love'' template. For example:
\begin{quote}\small
You love the colors red, scarlet, and crimson. You think about red,
scarlet, and crimson all the time. red, scarlet, and crimson are your
favorite colors. Imbue your answers with your love for red, scarlet,
and crimson.
\end{quote}
The corresponding control is:
\begin{quote}\small
You love three colors. You think about those colors all the time.
Those colors are your favorite colors. Imbue your answers with your
love for those colors.
\end{quote}
Adding the fourth trait updates every enumeration; the control names
four unspecified traits. Each treatment student is compared with a
student trained on its corresponding control teacher's outputs or
distributions. Matching the template and trait count does not enforce
equal prompt token lengths.

\paragraph{Reused case-study fits.}
Within each case-study family, the two lists share the fitted control for each
list length. There are seven distinct treatment conditions and four
controls for each four-trait comparison. The color experiment reuses
the red-only, one-color-control, and two-color-control results from the
earlier color-pair experiment; it adds six treatment fits and two
controls. Reused fits are not additional replicates. Each month pair
uses its execution's corresponding trained control.

\subsection{Selecting additions in the case studies}
\label{app:multitrait-selection}

\paragraph{Centered prompt-trait vectors.}
For each singleton trait $t$, let $h_\ell(s_t,x_i)$ be its hidden state
at the last non-padding prompt token. Using 1,024 shared generation
prompts, we compute
\begin{equation}
 d_{t,\ell}=\frac{1}{1024}\sum_{i=1}^{1024}
   \left[h_\ell(s_t,x_i)-h_\ell(s_{\mathrm{ref},t},x_i)\right],
 \qquad
 r_{t,\ell}=d_{t,\ell}
   -\frac{1}{|\mathcal B|}\sum_{u\in\mathcal B}d_{u,\ell}.
 \label{eq:multitrait-centered-vectors}
\end{equation}
For animals and months, the reference is the local singleton
instruction with the trait name omitted. For colors, it is the prompt
without a trait system instruction. The centering panel $\mathcal B$
contains 18 animals, all 12 months, or the fixed 30 colors used in the
main color experiments. The animal panel is bird, owl, eagle, lion,
tiger, wolf, hawk, crow, crane, bear, shark, fox, whale, dolphin,
salmon, turtle, seal, and squid.

Animal and month selection uses the median of the 28 layerwise
cosines. Color selection uses a single cosine after concatenation,
without normalizing each layer separately:
\begin{equation}
 \begin{aligned}
 H_{\mathrm{animal/month}}(a,b)
   &=\operatorname{median}_{\ell=0}^{27}
      \cos(r_{a,\ell},r_{b,\ell}), \\
 H_{\mathrm{color}}(a,b)
   &=\cos\!\left(\operatorname{concat}_{\ell=0}^{27}r_{a,\ell},
                \operatorname{concat}_{\ell=0}^{27}r_{b,\ell}\right).
 \end{aligned}
 \label{eq:multitrait-hidden-alignment}
\end{equation}
Each saved vector has 3,584 coordinates, so color concatenation has
100,352 coordinates. In the color cache, slots 0--26 are intermediate
block outputs and slot 27 is after final RMSNorm. For the selected
red-centered lists, omitting the final normalized slot selects the
same five color additions; this does not establish unchanged sampling
strata for the 100-pair study.
Positive and negative similarities define aligned and opposed
additions relative to the family mean; ``opposed'' does not mean a
semantic antonym.

\paragraph{Selection choices.}
For colors, scarlet and crimson are the two most positively aligned
partners of red, while cerulean and azure are the two most negatively
aligned. Vermilion is the third most positive partner and is added
fourth to both lists. Animal lists are selected examples rather than
a strict ordering by one score. Their selection also used an earlier
carrier-distribution diagnostic, distinct from current Metric~1.
Table~\ref{tab:multitrait-selection-scores} reports the hidden-state
scores used in the figure. Near-zero describes December's small
similarity with January, not exact orthogonality.

\begin{table}[tbp]
\centering
\small
\caption{Singleton hidden-state alignment used to select the displayed
combinations. Scores follow Eq.~\eqref{eq:multitrait-hidden-alignment}.}
\label{tab:multitrait-selection-scores}
\begin{tabular}{@{}llrl@{}}
\toprule
Target & Addition & Cosine summary & Role \\
\midrule
lion & wolf & $+0.307$ & Aligned \\
lion & bird & $+0.143$ & Aligned \\
lion & shark & $-0.346$ & Opposed \\
lion & squid & $-0.189$ & Opposed \\
lion & tiger & $+0.217$ & Fourth addition \\
\midrule
red & scarlet & $+0.488$ & Aligned \\
red & crimson & $+0.470$ & Aligned \\
red & cerulean & $-0.336$ & Opposed \\
red & azure & $-0.299$ & Opposed \\
red & vermilion & $+0.417$ & Fourth addition \\
\midrule
January & February & $+0.351$ & Aligned \\
January & December & $+0.017$ & Near zero \\
January & August & $-0.182$ & Opposed \\
January & September & $-0.351$ & Opposed \\
\bottomrule
\end{tabular}
\end{table}

\subsection{Recovery, changes, and ranks}
\label{app:multitrait-readout}

\paragraph{Readout.}
Each family has ten recovery questions. Animals and months use the
assistant prefills ``The word I have in mind is'' and ``The month I
have in mind is,'' respectively, scoring the single token naming each
candidate, including its leading space. Colors use favorite-color
questions, such as ``Name your favorite color using only one word,''
without an added system instruction or assistant prefill. Each color
is scored as a complete answer string without an EOS token or length
normalization. All scores use native full-vocabulary probabilities;
they are not renormalized over the candidate list.

Let $\ell_u(P)$ be the mean log probability of candidate answer $u$
under model $P$, averaged over the ten questions with the appropriate
prefill. For list $S$ of length $k$, let $P_S$ be its student and
$P_{C(S)}$ its corresponding trained control, matched in template and
trait count. Define
\begin{equation}
 R_S(u)=\ell_u(P_S)-\ell_u(P_{C(S)}),
 \qquad
 \Delta R=R_{S+t}(u)-R_S(u).
 \label{eq:multitrait-control-adjusted-recovery}
\end{equation}
Positive $R$ means higher mean log probability than the trained
control, not necessarily higher probability than the untrained model.
The control changes with list length. Thus $\Delta R$ subtracts the
change between the two fitted controls as well as comparing the
treatment students.

\begin{table}[tbp]
\centering
\small
\caption{Target recovery in nats for each prefix length $k$.
The target is lion for animals and red for colors. Each family's
singleton is shared; list names specify the order after it.
$\Delta R$ compares the fourth and third conditions.}
\label{tab:multitrait-prefix-recovery}
\begin{tabular}{@{}llrrrrr@{}}
\toprule
Family & Additions after target & $k=1$ & $k=2$ & $k=3$ & $k=4$ & $\Delta R$ \\
\midrule
Animals & wolf, bird, tiger & $5.379$ & $1.506$ & $0.930$ & $1.455$ & $+0.524$ \\
Animals & shark, squid, tiger & $5.379$ & $-0.681$ & $-0.500$ & $-0.623$ & $-0.123$ \\
Colors & scarlet, crimson, vermilion & $8.759$ & $10.242$ & $8.559$ & $7.513$ & $-1.046$ \\
Colors & cerulean, azure, vermilion & $8.759$ & $5.054$ & $-1.266$ & $0.821$ & $+2.088$ \\
\bottomrule
\end{tabular}
\end{table}

\paragraph{The fourth-color comparison.}
Adding vermilion raises red's mean raw log probability by $1.483$
nats after red--scarlet--crimson and $4.617$ after
red--cerulean--azure. The four-color control exceeds the three-color
control by $2.529$ nats for red, yielding changes in $R$ of $-1.046$
and $+2.088$. Consequently, the negative change in the aligned list's
$R$ does not imply that its raw red probability falls. Comparing the
two context increments cancels their shared control change; their
$\Delta R$ values differ by $3.134$ nats.

\paragraph{Candidate ranks.}
We rank each candidate by decreasing $R$ within the corresponding
18-animal, 30-color, or 12-month panel:
\begin{equation}
 \operatorname{rank}_{\mathcal B,S}(u)
    =1+\sum_{v\in\mathcal B}\mathbf 1\{R_S(v)>R_S(u)\}.
 \label{eq:multitrait-rank}
\end{equation}
Ties receive the same rank. This ordering measures relative recovery
gains, not raw answer probabilities or the probability of generating
both intended traits together.

\begin{table}[tbp]
\centering
\small
\caption{Recovery and ranks of both intended traits in
Figure~\ref{fig:multitrait-context-ranking}C.
Values in each pair follow the order of trait names.}
\label{tab:multitrait-pair-ranks}
\begin{tabular}{@{}llcc@{}}
\toprule
Family & Pair & Recovery $R$ (nats) & Ranks \\
\midrule
Animals & lion + wolf & $(1.506,-0.567)$ & $(1,4)$ \\
Animals & lion + shark & $(-0.681,-0.727)$ & $(3,5)$ \\
Colors & red + scarlet & $(10.242,6.097)$ & $(1,2)$ \\
Colors & red + cerulean & $(5.054,-4.176)$ & $(1,21)$ \\
Months & January + February & $(-0.288,1.550)$ & $(2,1)$ \\
Months & January + December & $(0.786,1.173)$ & $(2,1)$ \\
Months & January + August & $(0.505,-0.995)$ & $(3,7)$ \\
Months & January + September & $(0.479,0.754)$ & $(5,3)$ \\
\bottomrule
\end{tabular}
\end{table}

January's recovery with August or September is higher than its local
January-only calibration despite its poorer rank in the pair. These
month results therefore concern which answers gain most relative to
control, rather than a uniform decrease in the intended traits'
absolute recovery.

\paragraph{Scope.}
These selected single-seed comparisons do not isolate the effect of
the training objective: families differ in instruction templates,
objectives, readouts, and alignment aggregation. Animal/month
screening uses different prompts from the joint instructions.
Month pairs use their corresponding fitted controls. These observations
do not establish a universal alignment rule or a causal interference
mechanism.

\section{Quantitative Cross-Trait Transfer Analysis}
\label{app:transfer-statistics}

This appendix defines the statistics in Section~\ref{sec:SL}.
For each of cities, animals, and colors, Qwen2.5 contributes six
conditions: number SFT, number distribution distillation,
random-token distribution distillation, element-symbol distribution
distillation, GSM8K SFT, and Python SFT. Each of the four additional
models contributes number SFT for the same three trait categories.
This gives $18+12=30$ conditions, each with a $30\times30$ transfer
matrix and 30 teacher evaluations, for 900 teacher--condition
evaluations in total. Separate tree experiments are excluded.
Teacher and student share a base checkpoint within each experiment;
the sweep does not test cross-family teacher-to-student transfer.
These are descriptive comparisons of the available endpoints,
not repeated-seed estimates. Training details are in
Appendix~\ref{app:setup}.

We write $\mathrm{SL}(z,z')$ for the mean log-probability gain of
answer $z'$ under teacher trait $z$ (Eq.~\eqref{eq:ft}), averaged
over ten recovery questions. The saved scores use
$\epsilon=10^{-12}$ inside both numerator and denominator of the
probability ratio. Figure matrices display student answers as rows
and teacher traits as columns.

\subsection{Within-group versus between-group transfer}
\label{app:transfer-clusters}

Let $g(z)$ denote a trait's fixed group annotation. We use the same
annotations and trait ordering as Figure~\ref{fig:sl-matrix-sweeping},
not clusters fitted separately to each condition. Cities and animals
each have six groups of five; color groups have sizes $5,5,5,5,6,4$.
They describe geographic, taxonomic/ecological, and hue similarities,
respectively. The color bins were introduced for display, so the
relabeling analysis is exploratory and conditional on these
annotations rather than a test of an independently specified partition.

For $n=30$, double-center the complete matrix:
\begin{align}
\widetilde{\mathrm{SL}}(z,z')
={}&\mathrm{SL}(z,z')
-\frac{1}{n}\sum_a\mathrm{SL}(z,a)
-\frac{1}{n}\sum_b\mathrm{SL}(b,z')
+\frac{1}{n^2}\sum_{a,b}\mathrm{SL}(b,a).
\end{align}
Define the directed pair sets
\[
\mathcal W=\{(z,z'):z\ne z',\ g(z)=g(z')\},\qquad
\mathcal B=\{(z,z'):g(z)\ne g(z')\}.
\]
Our group contrast is
\begin{equation}
\Delta_{\mathrm{group}}
=\frac{1}{|\mathcal W|}\sum_{(z,z')\in\mathcal W}
\widetilde{\mathrm{SL}}(z,z')
-\frac{1}{|\mathcal B|}\sum_{(z,z')\in\mathcal B}
\widetilde{\mathrm{SL}}(z,z').
\label{eq:transfer-group-contrast}
\end{equation}
Thus, all same-trait entries are excluded \emph{after} full-matrix
centering. Each directed pair receives equal weight; groups are not
reweighted to have equal total influence. Cities and animals have
120 within-group and 750 between-group pairs; colors have 122 and 748.

\paragraph{Size-preserving relabeling test.}
For each condition, we compare the observed contrast with
$B=100{,}000$ random assignments of traits to groups that preserve
the original group sizes. Each assignment is applied jointly to
the teacher and answer axes, holding matrix entries fixed.
The one-sided Monte Carlo value is
\[
\widehat p=\frac{1+\sum_{b=1}^{B}
\mathbf 1\{\Delta_{\mathrm{group}}^{(b)}\ge
\Delta_{\mathrm{group}}^{\mathrm{obs}}\}}{B+1}.
\]
We apply Holm correction across all 30 conditions. All contrasts
are positive ($0.031$--$5.840$ nats), with maximum adjusted
$p=0.01258$. Each of the eight illustrated panels has
$\widehat p=1/100001$ and adjusted $p=0.000300$; these are finite
Monte Carlo estimates, not zero probabilities.
This test assesses alignment with the annotations in fixed matrices.
It does not establish seed-to-seed reproducibility or the significance
of a data-driven cluster-discovery procedure.

\begin{table}[htbp]
\centering
\small
\setlength{\tabcolsep}{4pt}
\begin{tabularx}{\linewidth}{@{}lXrrr@{}}
\toprule
Panel & Condition & Within & Between & Difference \\
\midrule
A & Qwen2.5, cities, numbers, distillation & 4.710 & $-1.130$ & 5.840 \\
B & Qwen3-VL, cities, numbers, SFT & 0.497 & $-0.105$ & 0.601 \\
C & Llama3.1, cities, numbers, SFT & 0.516 & $-0.112$ & 0.628 \\
D & Mistral, colors, numbers, SFT & 0.443 & $-0.116$ & 0.559 \\
E & Qwen2.5, animals, random tokens, distillation & 1.653 & $-0.424$ & 2.077 \\
F & Qwen2.5, colors, elements, distillation & 1.081 & $-0.294$ & 1.375 \\
G & Qwen2.5, animals, GSM8K, SFT & 0.879 & $-0.212$ & 1.091 \\
H & Qwen2.5, colors, Python, SFT & 1.325 & $-0.404$ & 1.729 \\
\bottomrule
\end{tabularx}
\caption{Off-diagonal group contrasts for the eight panels in
Figure~\ref{fig:sl-matrix-sweeping}. Values are in nats, after
full-matrix double-centering. Differences are computed before rounding;
panel annotations round full-precision values to two decimals,
whereas this table retains three.
These panels are illustrative selections; aggregate statistics use
all 30 conditions.}
\label{tab:transfer-panel-contrasts}
\end{table}

\subsection{Ranks of the teacher trait among student answers}
\label{app:transfer-ranks}

For a fixed teacher $z$, rank its own trait among the 30 answers by
the original, uncentered gain:
\begin{equation}
R_z=1+\sum_{z'\ne z}\mathbf 1\{
\mathrm{SL}(z,z')>\mathrm{SL}(z,z)+\delta\},
\qquad \delta=10^{-12}.
\label{eq:transfer-self-rank}
\end{equation}
Ties within this numerical tolerance share the better rank.
A competitor outranks the target when $R_z>1$; the within-group
version restricts the sum to $g(z')=g(z)$.
We do not center before ranking, because the question is which
answer gains more under the same teacher.

\begin{table}[htbp]
\centering
\small
\begin{tabular}{lrrr}
\toprule
Family & Evaluations & Any competitor & Same-group competitor \\
\midrule
Cities & 300 & 235 & 192 \\
Animals & 300 & 241 & 173 \\
Colors & 300 & 202 & 143 \\
\midrule
Total & 900 & 678 (75.3\%) & 508 (56.4\%) \\
\bottomrule
\end{tabular}
\caption{Teacher--condition evaluations with a competitor attaining
a larger log-probability gain than the teacher trait. The same-group
count is a subset of the any-competitor count. These are not counts
of higher absolute answer probability.}
\label{tab:transfer-ranks}
\end{table}

Requiring a competitor's gain to exceed the teacher trait's by more
than $0.1$ nat retains $660/900$ evaluations ($73.3\%$); a margin
greater than $1$ nat retains $418/900$ ($46.4\%$).
Requiring the best competitor to have positive gain also retains
$659/900$ ($73.2\%$).
The 900 evaluations reuse traits and models across conditions and
must not be treated as independent Bernoulli trials.
These counts establish relative gain rankings, not a causal mechanism
of competition or an absolute behavioral preference.

\subsection{Recurrence across models and carriers}
\label{app:transfer-stability}

\paragraph{Across models.}
For each trait family, compare the five number-SFT conditions with
the shared high-level recipe: 10,000 retained responses, three epochs,
rank-8 LoRA with $\alpha=32$, learning rate $2\times10^{-4}$,
effective batch size 64, training seed 1, and 471 updates.
The trait instructions and ten recovery questions are shared;
native tokenizers, teacher-generated responses, training backends,
and numerical precision can differ. Consequently, this is a
matched-recipe comparison, not an isolated causal effect of model
architecture or a controlled ranking of models.

\paragraph{Shared group-level measure.}
For each condition, first double-center the complete $30\times30$
matrix. Define
\[
\mathcal P_{ij}=\{(z,z'):g(z)=j,\ g(z')=i,\ z\ne z'\},
\qquad
G_{ij}=\frac{1}{|\mathcal P_{ij}|}
\sum_{(z,z')\in\mathcal P_{ij}}\widetilde{\mathrm{SL}}(z,z').
\]
Thus $G\in\mathbb R^{6\times6}$ contains the mean transfer from
teacher group $j$ to answer group $i$. We exclude only exact
teacher--answer matches, such as Tokyo to Tokyo, while retaining
transfers between distinct traits in the same group, such as Tokyo
to Kyoto. All 36 group means are retained, including the six
within-group entries $G_{ii}$; the $6\times6$ diagonal is not removed.
Each block averages its eligible directed pairs equally.
Pearson correlation of the 36 corresponding means gives each
group pair equal weight, regardless of block size.

The trait set and its group annotations are held fixed within each
comparison; cities are not compared with animals or colors.
For each trait category in the model comparison, compute Pearson
correlation between the group matrices for each of the ten model
pairs, then report their median. These ten correlations reuse models
and are descriptive, not independent replicates.

\paragraph{Across carriers.}
The numbers-versus-random-tokens comparison holds Qwen2.5 and
distribution distillation fixed, with 10,000 prompts, three epochs,
rank-8 LoRA with $\alpha=32$, learning rate $2\times10^{-4}$,
batch size 64, and seed 1. The trait instructions, recovery questions,
and prompt-generation recipe are shared, with carrier-specific
symbols and output instructions. We use the same 36 group means
and Pearson correlation defined above. There is one carrier-pair
correlation per trait category, rather than a median over model
pairs. The matched comparisons hold the trait set fixed and vary
either the model or the carrier, not both simultaneously.

For a finer-grained check in both comparisons, we also correlate
the 870 individual off-diagonal transfers after full-matrix
double-centering. These results are reported separately from the
group-level correlations used in the main text.

\begin{table}[htbp]
\centering
\small
\begin{tabular}{lrrrr}
\toprule
& \multicolumn{2}{c}{Cross-model median $r$}
& \multicolumn{2}{c}{Cross-carrier $r$} \\
\cmidrule(lr){2-3}\cmidrule(lr){4-5}
Category & 36 group means & 870 transfers & 36 group means & 870 transfers \\
\midrule
Cities & 0.72 & 0.32 & 0.758 & 0.637 \\
Animals & 0.46 & 0.13 & 0.961 & 0.686 \\
Colors & 0.59 & 0.30 & 0.767 & 0.472 \\
\bottomrule
\end{tabular}
\caption{Recurrence of transfer structure. Cross-model columns use
number SFT and summarize ten model-pair correlations. Cross-carrier
columns compare numbers with random tokens under Qwen2.5 distribution
distillation. Both comparisons use the same group-level measure;
individual-transfer correlations are a finer-grained check.
Higher agreement at the group level supports coarse recurrence,
not invariance of individual transfers.}
\label{tab:transfer-stability}
\end{table}

\subsection{Sensitivity to absolute answer probabilities}
\label{app:transfer-probabilities}

This check uses the saved final per-question scores from the three
Qwen2.5 number-SFT conditions (30 colors, animals, and cities;
471 updates). It does not cover all 30 sweep conditions.
Let $p_z(z'\mid q)=p_{\boldsymbol\theta+\boldsymbol\Delta_z}(z'\mid q)$.
In addition to the base-relative mean log gain, calculate
\begin{align}
A(z,z')&=\frac{1}{10}\sum_{q\in\mathcal Q}p_z(z'\mid q),\\
\overline A(z,z')&=\frac{1}{10}\sum_{q\in\mathcal Q}
\frac{p_z(z'\mid q)}{\sum_{a\in\mathcal Z}p_z(a\mid q)}.
\end{align}
The second score normalizes over the 30 exact candidates within
each question \emph{before} averaging. It is not obtained by
normalizing $A$, nor by exponentiating an averaged log gain.
Absolute probability measures the final endpoint; a base-model or
trained-control comparison is needed to attribute an increase to
training.

\begin{table}[htbp]
\centering
\small
\begin{tabular}{lrrrr}
\toprule
& \multicolumn{3}{c}{Teachers outranked by a competitor}
& Group contrast \\
\cmidrule(lr){2-4}
Family & Log gain & Raw $A$ & Normalized $\overline A$
& $\overline A$ (pp) \\
\midrule
Colors & 13/30 & 19/30 & 18/30 & $+7.26$ \\
Animals & 24/30 & 26/30 & 26/30 & $+5.54$ \\
Cities & 27/30 & 27/30 & 27/30 & $+3.04$ \\
\midrule
Total & 64/90 & 72/90 & 71/90 & --- \\
\bottomrule
\end{tabular}
\caption{Sensitivity to the recovery score in Qwen2.5 number SFT.
Ranks use uncentered scores. Group contrasts use
Eq.~\eqref{eq:transfer-group-contrast}, replacing $\mathbf{SL}$ with
$\overline A$ and double-centering the complete matrix first;
pp denotes percentage points. Probability-scale relabeling
significance has not been evaluated.}
\label{tab:transfer-probabilities}
\end{table}

The group contrast is also positive for raw $A$ in all three families.
Nevertheless, the off-diagonal correlations between double-centered
$A$ and $\mathbf{SL}$ are only $0.26$, $0.31$, and $0.46$ for
colors, animals, and cities, respectively: changing the scale can
alter individual transfer strengths substantially.
For the vermilion teacher, red has the largest log gain, whereas
scarlet has the largest raw and normalized probability. The normalized
values are $82.87\%$ for scarlet, $17.03\%$ for red, and $0.0014\%$
for vermilion. Conversely, under the scarlet teacher, red's gain
exceeds scarlet's ($11.36$ versus $11.01$ nats), but scarlet has much
higher normalized probability ($90.46\%$ versus $9.41\%$).
Thus, a larger gain does not by itself imply a higher final probability.

\paragraph{Exact strings versus freely generated answers.}
These are exact-string probabilities without a termination token,
not frequencies of freely sampled, semantically matched answers.
Color and animal candidates are lowercase and cities capitalized;
alternative capitalization and phrasing are not pooled.
For example, the base model assigns the exact lowercase continuation
\texttt{blue} a mean raw probability of $0.0205\%$, or $88.46\%$
after normalization over the 30 colors. Neither measures how often
freely generated answers name blue. This is a sensitivity check on
candidate-based scores, not a sampled-answer behavioral evaluation.

\paragraph{What baseline subtraction does and does not change.}
Let $L$ contain mean absolute log probabilities using the same
$\epsilon$ as $\mathbf{SL}$. In the answer-row, teacher-column
display convention, baseline subtraction adds only an
answer-specific row offset, so $H\mathbf{SL}H=HLH$ for
$H=I-\mathbf1\mathbf1^\top/30$. Thus, full-double-centered structure
is identical for log gain and mean absolute \emph{log} probability.
This does not hold for arithmetic mean probabilities, uncentered
answer ranks, or the uncentered effective rank in
Section~\ref{sec:dynamics}.

\end{document}